\documentclass[10pt,twocolumn]{article}

\usepackage[margin=0.75in,columnsep=0.3in]{geometry}
\usepackage{graphicx}
\usepackage{multirow}
\usepackage{amsmath,amssymb,amsfonts}
\usepackage{xcolor}
\usepackage{textcomp}
\usepackage{booktabs}
\usepackage{algorithm}
\usepackage{algorithmicx}
\usepackage{algpseudocode}
\usepackage{hyperref}
\usepackage{flushend}
\usepackage{caption}
\usepackage{titlesec}
\titleformat{\section}{\normalfont\large\bfseries}{\thesection}{1em}{}
\titleformat{\subsection}{\normalfont\bfseries}{\thesubsection}{1em}{}

\title{\LARGE Neuro-Geospatial Modelling of EEG Affective States Using Literature-Informed Environmental Context}

\author{
Utsav Poudel$^{1}$, \; Jagannath Aryal$^{2}$, \; Subramaniyaswamy Vairavasundaram$^{1}$\\[6pt]
\small $^{1}$School of Computer Science and Engineering, Vellore Institute of Technology,\\
\small Vellore, Tamil Nadu 632014, India\\[2pt]
\small $^{2}$Earth Observation and AI Research Group, Department of Infrastructure Engineering,\\
\small The University of Melbourne, Melbourne, VIC 3010, Australia
}
\date{}

\begin{document}

\twocolumn[
\begin{@twocolumnfalse}
\maketitle
\begin{abstract}
\noindent Environmental exposures such as air pollution and greenness are linked to affective and cognitive outcomes, but EEG and environmental datasets are rarely jointly georeferenced, limiting direct evaluation. We investigate whether literature-informed environmental priors can be incorporated as an auxiliary geospatial modality for EEG-based affective-state classification when individual-level environmental exposure is unavailable. We combine 30-channel EEG from the EAV benchmark (42 participants, aged 20--30) with environmental representations derived from OpenAQ, Sentinel-2, Sentinel-5P and OpenStreetMap data for Astana. A dual-tower architecture combines EEG-Conformer representations with a graph-based environmental encoder. Because the datasets were not co-registered, environmental context is a literature-informed prior rather than measured exposure. Subject-level repeated splits, permutation/label-shuffling controls, dose-response reversal, and domain-shift experiments separate architecture-level gains from prior-dependent gains. The multimodal model reaches $76.2$\% accuracy versus $67.4$\% for EEG alone; controls disrupting environmental-label structure retain part of this gain, so it is not attributable solely to environmental information. Substituting an independently modelled Singapore environmental distribution for Astana's reduces accuracy to $72.8$\%. The results support the technical feasibility of this approach but do not establish an observed or causal exposure-affect association. The study provides a methodological framework for future jointly collected mobile EEG-environment studies. Implementation at: \url{https://github.com/r11up/geo-cog}.
\end{abstract}
\vspace{4pt}
\vspace{14pt}
\end{@twocolumnfalse}
]

\section{Background}\label{sec:background}

Urban environmental conditions, air pollution, traffic density, limited access to vegetation, have repeatedly been associated with variation in psychological and cognitive outcomes in epidemiological and environmental-health research \cite{Rodriguez2026}. Fine particulate matter (PM$_{2.5}$) exposure is linked to reduced executive function ($\beta=-0.0146$, $p<.001$) \cite{Yu2026} and, over longer exposure windows, to faster cognitive decline \cite{Diao2026,Faherty2025}; urban noise is tied to psychological stress and disrupted sleep \cite{noise2024}; access to green space is tied to lower stress and better mood \cite{Wang2026a,Forestry26}. This evidence provides a credible basis for hypothesising that environmental context is relevant to affective and neurophysiological states, but it says little about how that context operates moment to moment: it is built almost entirely from population-level exposure indicators and long-term aggregate outcomes rather than temporally resolved neural recordings.

Electroencephalography (EEG) is a natural complement to this literature. It records cortical activity at millisecond resolution, sensitive to fluctuations in arousal, valence and cognitive load that self-report and scanner-based neuroimaging cannot capture outside a laboratory, the temporal sensitivity that would eventually be needed to couple neural responses with fast-changing, spatially heterogeneous urban stimuli. Geospatial cognition research proposes that the brain continually encodes and updates spatial representations of its surroundings to guide behaviour \cite{main_paper,Han2025,Langyuan}, and a related literature has begun asking how elements such as air pollution, density and visual complexity function as spatially distributed influences on brain-mediated behaviour and mental-health outcomes \cite{green_gentrification,Falkenstein26,Vulnerability,greenspacemeta2023}. EEG-based affective-state research, though, has been almost entirely confined to controlled laboratory settings disconnected from real-world environmental context \cite{brain_Marino,Almuntasheri26}, and few studies have attempted to combine the two directly \cite{WANG26}.

This is a specific methodological problem rather than a simple absence of evidence. Environmental datasets, of the kind used here, are spatially detailed and available at no cost in almost any city. EEG affective-state benchmarks are temporally detailed but collected in a laboratory, with no simultaneous record of a participant's outdoor location or exposure. The EAV dataset used in this study \cite{lee2024eav} is no exception: participants wore EEG equipment at Nazarbayev University in Astana, Kazakhstan, and no participant-level trajectory, outdoor sensor reading, or timestamped exposure measurement exists alongside their recordings. Any environmental vector assigned to an EEG epoch in this study is therefore not that participant's measured exposure at the time of recording; it is a value sampled from the city-wide environmental distribution and, during training only, conditioned on the epoch's affective label according to dose-response relationships reported in the epidemiological literature (Methods). Because of this, the present study cannot estimate a relationship between exposure and affect empirically. What it can do is ask a narrower question: can a multimodal deep-learning architecture learn to exploit a literature-consistent, spatially structured environmental prior when that prior is deliberately built into training, and does the resulting benefit survive label-independent evaluation and a change of city?

Three boundaries follow directly from this and are worth stating before the methodological detail. First, the environmental vector attached to any EEG epoch here is a computational construction, not a measured exposure. Second, the framework is accordingly a computational digital twin, a controlled simulation of environmental context layered onto a laboratory EEG benchmark, not a system that records neurophysiological responses to real outdoor stimuli. Third, every reported association, including the Singapore evaluation, is evidence about model and architecture behaviour, not about a causal or directly observational coupling between brain and environment; that requires temporally and spatially co-registered in-situ data, which we return to in Discussion as future work.

Environmental context is conventionally represented through geographic interpolation, buffer analysis, or land-use regression (LUR) \cite{Rodriguez2026,lur}. LUR relates ground-level pollutant measurements to spatial predictors such as road density and building morphology, and sits between pure geostatistical interpolation, which cannot capture the sharp, morphology-driven pollution gradients typical of dense urban areas, and dispersion or chemical-transport models, which need emissions and meteorological inputs rarely available outside well-instrumented cities \cite{Hennig2016}. Green-space access, measured here through Sentinel-2 vegetation indices, is repeatedly linked to reduced depression and anxiety \cite{Large-scale,Vulnerability}, and OpenStreetMap-derived morphology metrics, Floor Area Ratio (FAR) and a satellite-derived vegetation-visibility proxy, capture density and visual exposure to nature \cite{Wang2026a}. Mechanistically, green-space exposure is thought to lower cortisol and attentional fatigue through pathways described by Attention Restoration Theory and Stress Reduction Theory \cite{Forestry26,li2026}, while high-density built form and traffic pollution are linked to elevated stress biomarkers \cite{Mueller2020}, and functional MRI work ties urban green infrastructure to prefrontal and amygdala activity involved in emotion regulation \cite{Jiang2023}.

On the neural side, spectral features, particularly Differential Entropy (DE) and Power Spectral Density (PSD) across the $\delta$, $\theta$, $\alpha$, $\beta$ and $\gamma$ bands, carry discriminative information for valence, arousal and cognitive load in benchmark EEG datasets \cite{Koelstra2012,Liu2026}, and hybrid decoders combining convolutional and transformer components such as the EEG-Conformer \cite{Song2023} now outperform earlier convolutional-only architectures \cite{Paredes2025}. Multimodal fusion of EEG with other biosignals typically uses unidirectional cross-modal attention, treating one modality as query and the other as context \cite{multieeg,Lee2024}; bidirectional attention, where each modality serves as both query and context for the other, has been used for tasks combining speech and vision but not previously for EEG and geographic data. Because EEG covariance matrices are symmetric positive definite and lie on a curved manifold, Riemannian-geometry alignment has outperformed Euclidean domain adaptation in brain-computer-interface work \cite{barachant2012}, which partly motivated our alignment design (Methods). Spatiotemporal graph neural networks encode Tobler's first law of geography, that near things are more related than distant things, as a structural inductive bias \cite{tobler1970,Xiao2024}, extending earlier ontology-driven GIScience approaches to multimodal spatial prediction \cite{GIScience266,Rajbhandari2019}.

This study addresses the co-registration problem above with a bidirectional neuro-geospatial architecture, an EEG-Conformer paired with a graph neural network over a 100-cell environmental grid, coupled by bidirectional cross-modal attention, and a two-stage protocol for linking the two data domains computationally. The contributions of this work are as follows. 1. A framework for integrating unco-registered EEG and geospatial environmental data through a literature-informed prior (PECM) and a directly optimised cross-modal alignment objective. 2. A controlled protocol for distinguishing the EEG signal, the environmental auxiliary information, and the architectural multimodal benefit from one another. 3. An explicit evaluation of how the environmental prior behaves under environmental-domain shift, re-fitting only the environmental tower to an independently modelled Singapore distribution while the EEG encoder and its Astana training are held fixed. 4. A reproducible, openly available pipeline that provides a pathway toward future studies pairing mobile EEG with georeferenced exposure measurements.

We treat this two-city design, rather than a single deep case study, as central to the paper's contribution. Every environmental input used here, ground-station data from OpenAQ, building and road vectors from OpenStreetMap, atmospheric imagery from the Sentinel missions, is available at no cost in essentially any city, including many with sparse dedicated air-quality monitoring. The Singapore evaluation tests whether the same architecture, re-fitted only at the environmental tower, keeps working when the underlying exposure distribution is replaced by an independently modelled city, the property a researcher elsewhere would actually need to reuse this template.

\subsection{Research questions and hypotheses}\label{subsec:rq}

1. Does incorporating a literature-informed environmental context improve EEG-based affective-state classification relative to an EEG-only baseline, and how much of that improvement reflects the environmental content itself rather than the added architecture? 2. Does the environmental context remain useful for classification when the environmental domain shifts to a structurally different, independently modelled city? 3. How sensitive are the results to spatial scale and to the construction, direction and resolution of the environmental context?

Correspondingly: a fusion model using the literature-informed context should outperform an EEG-only baseline, with part of that improvement persisting under label-decorrelated controls, an architecture-level contribution, and part attributable to the literature-informed content specifically (H1); environmental-domain substitution should reduce, but not eliminate, the multimodal advantage (H2); and performance should depend on the spatial resolution of the environmental context and on the literature-consistent direction of the injected association, not merely on the presence of some partition (H3). RQ1 is addressed by the baseline-versus-proposed-model and ablation comparisons (Sections~\ref{subsec:mainresults} and \ref{subsec:ablation}), RQ2 by the Singapore environmental-domain shift evaluation (Section~\ref{subsec:shift}), and RQ3 by the spatial-scale and dose-response-direction sensitivity analyses (Sections~\ref{subsec:maup} and \ref{subsec:ablation}). These are hypotheses about computational performance and robustness, not about a causal effect of environmental exposure on any individual's neural state.

\section{Data and study design}\label{sec:methods}

\subsection{Study design overview}\label{subsec:design}

\begin{figure*}[t]
\centering
\includegraphics[width=0.88\textwidth]{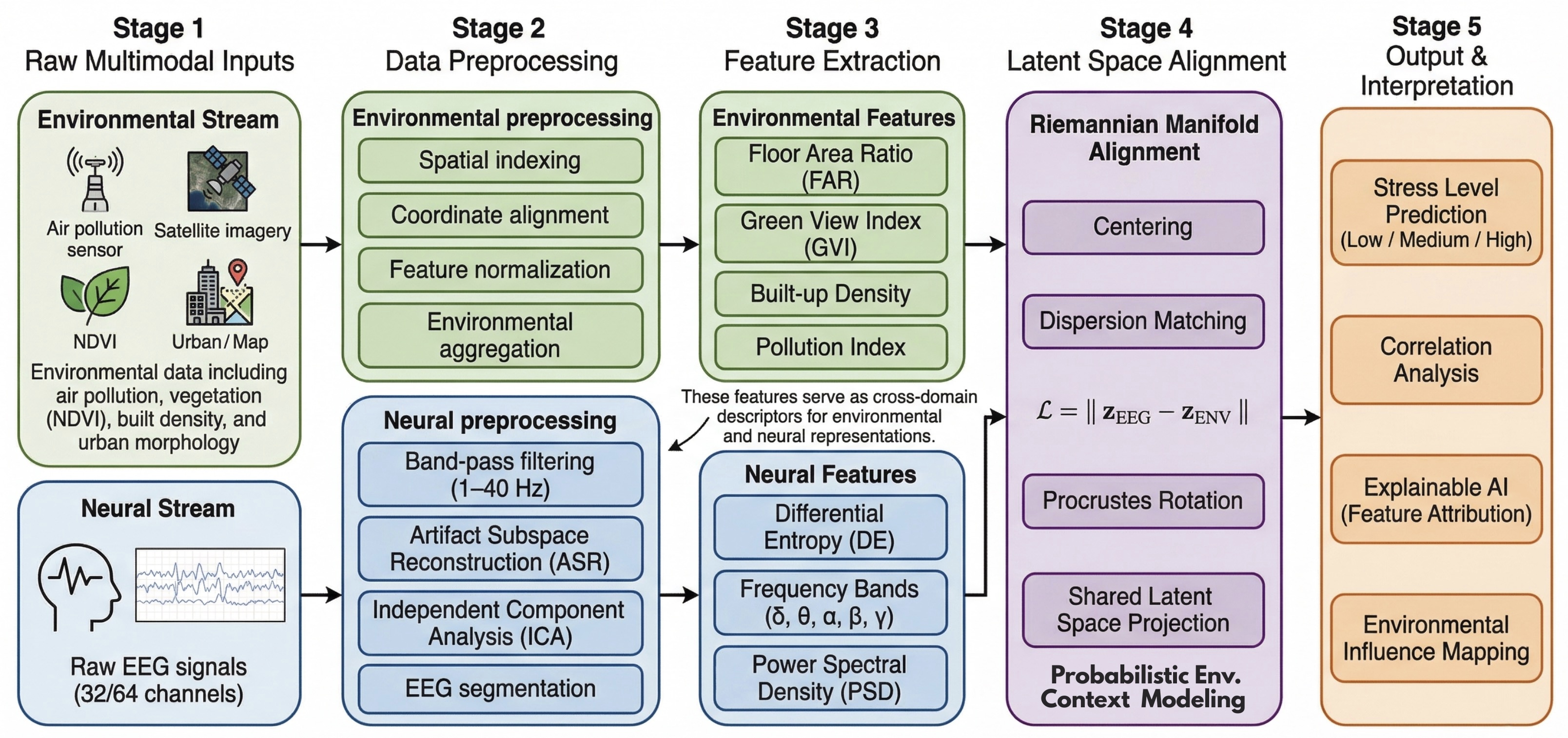}
\caption{Data-integration pipeline. EEG recordings and environmental context are linked through PECM and aligned via a directly optimised squared-$\ell_2$ objective (Supplementary S2, S6).}
\label{fig:framework}
\end{figure*}

This study combines one outcome, the momentary affective-state label attached to each EEG epoch, with two predictor families: the EEG signal itself, and multi-source geospatial data describing morphological, ecological and atmospheric conditions across two urban study areas. Because the neural and environmental domains were not co-registered in real time, a two-stage protocol links them computationally. Probabilistic Environmental Context Modeling (PECM) first establishes label-conditioned correspondence based on epidemiological dose-response evidence; a shared representation is then learned through a directly optimised cross-modal alignment objective, motivated by, and validated against, a Riemannian-geometry formulation. PECM should be read throughout as a literature-informed prior-generation mechanism, not as a reconstruction of any participant's actual exposure. For the primary Astana analysis, the environmental data are drawn from the EEG participants' own city, which anchors PECM's conditioning in genuine population-level geographic co-location, though not in individual-level spatial co-registration (Discussion). The full data-integration pipeline is summarised in Figure~\ref{fig:framework}, and the complete mathematical specification of every model component, including the full Riemannian alignment algorithm, is given in Supplementary Material S2 and S6.

\subsection{Participants and neurophysiological data}\label{subsec:participants}

The outcome variable and neural predictor are both drawn from the EAV (EEG-Audio-Video) benchmark \cite{lee2024eav}, a resource of 30-channel EEG from 42 participants at a native sampling rate of 500~Hz, recorded at Nazarbayev University in Astana, Kazakhstan and available upon authorised request from the original creators. This gives population-level geographic co-location between the EEG participants and the primary environmental study area; Discussion addresses why that is not the same as individual-level exposure co-registration. Participants were adults; Lee et al.\ \cite{lee2024eav} report an age range of 20--30 years for this cohort, and full recruitment and demographic details, including gender breakdown, are available in the original publication. There were 20 trials per condition, each 20 seconds long (200 trials per participant). The five \emph{speaking} conditions used here, neutral, anger, sadness, happiness, relaxed/calm, involve active, behaviourally engaged affective states, closer to the real-world scenarios of interest than the passive-listening conditions in the same dataset.

The label attached to each trial is the intended, elicited emotion for its scripted condition, not a freely reported or spontaneously occurring one: participants selected the dialogue scenario they found most evocative for each target category, then alternately listened to and spoke pre-scripted dialogue designed to elicit it, with arousal and valence additionally rated by participants and experimenters \cite{lee2024eav}.

Each participant's raw EEG is a tensor of shape $(10{,}000 \times 30 \times 200)$ (10,000 samples per trial at 500~Hz, 30 electrodes, 200 trials), of which the five speaking conditions contribute $5\times20=100$ trials. Signals were band-pass filtered, artifact-corrected (Artifact Subspace Reconstruction \cite{Miyakoshi2023}, FastICA \cite{Kumaravel2023}) and segmented into four non-overlapping 5-second epochs per trial, 400 epochs per participant, 16,800 epochs in total; full filtering, downsampling and segmentation parameters are given in Supplementary Material S1. EAV was chosen over other benchmarks for its 30-channel montage, consistent with evidence that electrode count contributes more to EEG emotion-recognition accuracy than spatial topology alone \cite{Zhao2026b}, its five-class coverage of the spectrum from stress to restoration, and its balanced class distribution.

Each 5-second epoch yields Differential Entropy (DE) and Power Spectral Density (PSD) features across five canonical frequency bands ($\delta$, $\theta$, $\alpha$, $\beta$, $\gamma$); DE is preferred over raw spectral power for its more linearly separable classes \cite{Goshvarpour2024}. Concatenating DE and PSD across all 30 channels and five bands gives the neural input tensor $\mathbf{X}\in\mathbb{R}^{T\times30\times10}$ consumed by the neural tower (Section~\ref{subsec:architecture}); full computation parameters and the differential-entropy derivation are given in Supplementary Material S1.

\subsection{Study areas}\label{subsec:studyareas}

\begin{figure*}[t]
\centering
\includegraphics[width=0.88\textwidth]{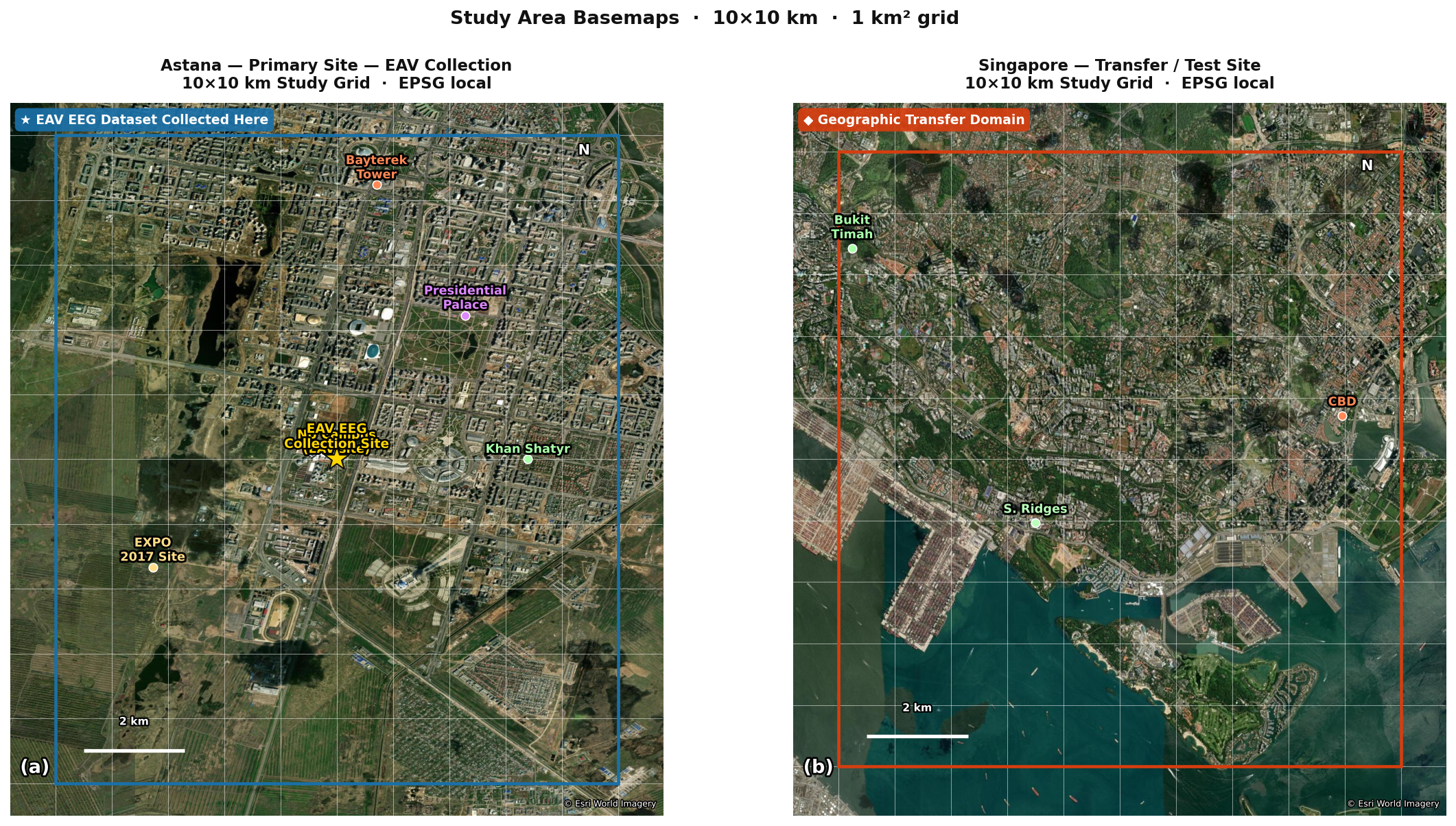}
\caption{Study area base maps, $10\times10$~km grids for Astana (primary site; EAV EEG collected here) and Singapore (secondary site, environmental-domain shift evaluation only; no Singapore EEG data used).}
\label{fig:studyarea}
\end{figure*}

\begin{figure*}[t]
\centering
\includegraphics[width=0.9\textwidth]{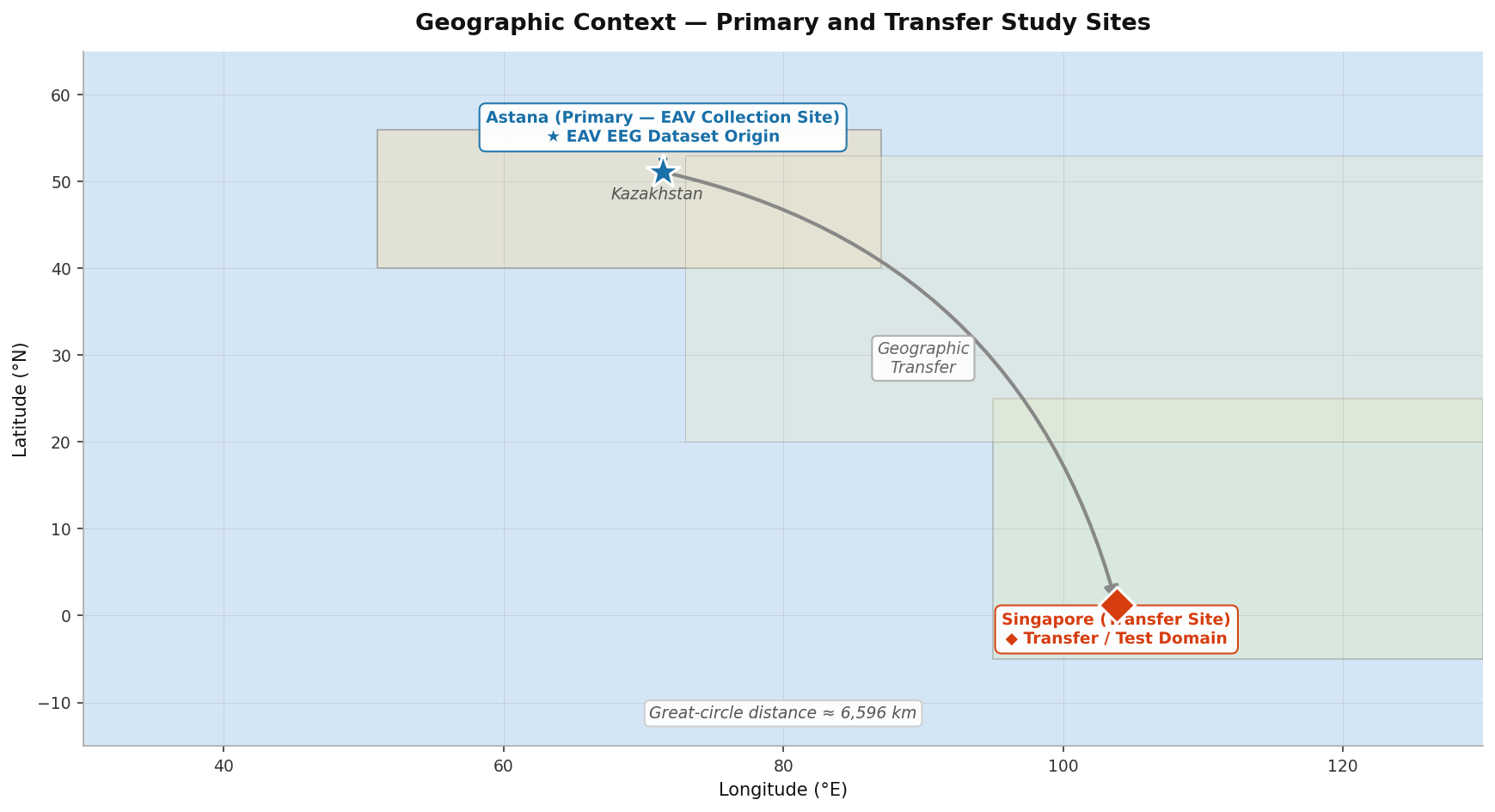}
\caption{Geographic context for the environmental-domain shift evaluation: Astana (EEG collection site) and Singapore (independently modelled environmental distribution only; no Singapore EEG data used), separated by 6,596~km, confirming the two environmental domains are structurally independent.}
\label{fig:transfer}
\end{figure*}

The primary geospatial domain covers a $10\times10$~km area centred on Nazarbayev University in Astana, Kazakhstan ($51^{\circ}53'$N, $71^{\circ}26'$E), projected to UTM Zone~41N (EPSG:32641) and discretised into a $1~\text{km}^2$ grid of 100 spatial units. Astana was selected as the primary study area because it is genuinely the EEG participants' daily urban environment at the population level, has a continental climate with well-documented seasonal PM$_{2.5}$ elevation from traffic and district heating, and hosts operational OpenAQ stations that anchor the LUR model. Singapore's central planning area ($1^{\circ}17'$N, $103^{\circ}49'$E), projected to the Singapore Transverse Mercator system (EPSG:3414), serves as a secondary, structurally different environment for testing environmental-domain robustness, using the same $10\times10$~km, $1~\text{km}^2$ grid methodology. No EAV participants are Singapore residents, so this evaluation runs the full PECM protocol without Astana's co-location advantage; the deliberate contrast in building density and seasonal vegetation between the two cities is what makes the comparison informative (Figures~\ref{fig:studyarea} and \ref{fig:transfer}).

\subsection{Environmental context modelling: land use regression}\label{subsec:lur}

Environmental data are assembled from four globally available platforms spanning atmospheric, ecological and morphological dimensions of the urban environment (Table~\ref{tab:datasets}); identical sources and processing are applied to both study areas, which is what makes the pipeline reusable elsewhere. Atmospheric pollution is estimated with an LUR model combining OpenAQ ground measurements of PM$_{2.5}$ and NO$_2$ with spatial predictors: road density, building Floor Area Ratio (FAR), vegetation (NDVI), distance to major roads, and Sentinel-5P TROPOMI data used solely as a coarse background predictor. For each pollutant $p\in\{\mathrm{PM_{2.5}},\mathrm{NO_2}\}$:
\begin{align}
\hat{c}_{p}(\mathbf{x}) &= \beta_0^p + \beta_1^p\,\rho_{\mathrm{road}}(\mathbf{x}) + \beta_2^p\,\mathrm{FAR}(\mathbf{x}) \notag\\
&\quad + \beta_3^p\,c_p^{\mathrm{TROPOMI}}(\mathbf{x}) + \beta_4^p\,\mathrm{NDVI}(\mathbf{x}) \notag\\
&\quad + \beta_5^p\,d_{\mathrm{road}}(\mathbf{x}) + \varepsilon(\mathbf{x})
\label{eq:lur}
\end{align}
where $\rho_{\mathrm{road}}$ is road length density and $d_{\mathrm{road}}$ is distance to the nearest arterial road. Coefficients are estimated by ordinary least squares with leave-one-station-out cross-validation. For Astana, calibrated on $N_A=6$ OpenAQ stations, the LUR reaches $R^2=0.74$ for PM$_{2.5}$ and $R^2=0.68$ for NO$_2$; for Singapore, calibrated on $N_S=9$ stations, $R^2=0.71$ and $0.65$. Six and nine calibration stations are a genuine limitation on the spatial representativeness of the resulting surfaces; full regression diagnostics (variance inflation factors, residual normality, spatial autocorrelation) and the rationale for preferring a sparse-network LUR over coarse satellite retrievals in lower-monitoring-density settings are given in Supplementary Material S4.

Vegetation cover uses Sentinel-2 Level-2A imagery \cite{sentinel2}, a cloud-masked annual composite at 10~m resolution:
\begin{equation}
\mathrm{NDVI} = \frac{B_8-B_4}{B_8+B_4}
\label{eq:ndvi}
\end{equation}
where $B_8$ and $B_4$ are near-infrared and red reflectance; cells with NDVI$>0.3$ are categorised as vegetated \cite{Large-scale,Aryal2022}. Urban morphology, road networks, floor-plate areas and building footprints, is taken from OpenStreetMap \cite{osm2017} and rasterised to the $1~\text{km}^2$ grid to calculate FAR; full multi-resolution processing and harmonisation of all four data layers (TROPOMI, LUR, Sentinel-2, OpenStreetMap) is given in Supplementary Material S4. The O$_3$ layer is retained at native Sentinel-5P resolution and is consequently smoother across neighbouring cells than the other node features, which matters when interpreting the feature-attribution results in Results.

\begin{table*}[t]
\caption{Datasets used in the multimodal neuro-geospatial framework.}\label{tab:datasets}
\begin{tabular}{@{}p{2.6cm}p{2.8cm}p{2.4cm}p{3.4cm}p{1.7cm}p{1.6cm}@{}}
\toprule
Dataset & Source & Modality & Key attributes & Spatial res. & Temporal res. \\
\midrule
EAV \cite{lee2024eav} & Lab EEG (Astana) & Neural & 30-ch, 42 participants, 5 affective states & --- & 500~Hz \\
OpenAQ \cite{opeanaq} & API & Air quality & PM$_{2.5}$, NO$_2$ & Point & Hourly \\
Sentinel-5P \cite{sentinel5p} & Satellite & Atmospheric & NO$_2$, O$_3$, aerosol & 3.5~km$^\dagger$ & Daily \\
LUR model & OpenAQ+OSM+S5P & Atmospheric & PM$_{2.5}$, NO$_2$ & 1~km$^2$ & Annual \\
Sentinel-2 \cite{sentinel2} & Satellite & Vegetation & NDVI & 10~m & 5-day \\
OpenStreetMap \cite{osm2017} & Open vector & Urban morphology & Buildings, roads, FAR & Parcel & Static \\
\bottomrule
\end{tabular}
\begin{flushleft}
\footnotesize $^\dagger$TROPOMI is used only as a background predictor (Eq.~\eqref{eq:lur}). Astana: EPSG:32641; Singapore: EPSG:3414. All sources freely available.
\end{flushleft}
\end{table*}

\subsection{Environmental indicators and node features}\label{subsec:indicators}

Three composite indicators reflect the environmental factors most consistently linked to neurophysiological outcomes in the literature used to construct PECM: LUR-estimated PM$_{2.5}$ \cite{Diao2026,Faherty2025}; FAR, a proxy for urban canyon morphology \cite{Jiang2023}; and a satellite-derived vegetation-visibility proxy (GVI) \cite{Li22021a}. NO$_2$ \cite{no2biomarkers2023} and O$_3$ enter the node feature vector directly. All indicators were standardised within each city to zero mean and unit variance. Indicators were computed at a 250~m buffer, which produced the most stable distributions and best matched pedestrian-scale ranges reported in environmental psychology \cite{li2026,Forestry26}; the full 100/250/500~m sensitivity comparison is reported in Supplementary Material S5.

Environmental features are arranged as a spatial graph $\mathcal{G}=(\mathcal{V},\mathcal{E},\mathbf{F})$ over the 100-node study grid, with a six-dimensional feature vector at each node $v_i\in\mathcal{V}$:
\begin{equation}
\resizebox{0.98\linewidth}{!}{$\displaystyle
\mathbf{f}_i = \bigl[\hat{c}_{\mathrm{PM_{2.5}}}(\mathbf{x}_i),\ \mathrm{NDVI},\ \mathrm{FAR},\ \mathrm{GVI},\ \hat{c}_{\mathrm{NO_2}}(\mathbf{x}_i),\ c_{\mathrm{O_3}}(\mathbf{x}_i)\bigr]^{\!\top}\in\mathbb{R}^6
$}
\label{eq:nodefeature}
\end{equation}
Edges $\mathcal{E}$ join spatially close nodes using queen's contiguity with inverse-squared-distance weights $w_{ij}=1/(d_{ij}^2+\epsilon)$, operationalising Tobler's first law of geography as a structural inductive bias \cite{tobler1970}.

\subsection{Cross-domain integration: PECM as a literature-informed prior}\label{subsec:alignment}

Because the neural and environmental domains were not co-registered in real time, we use Probabilistic Environmental Context Modeling (PECM) to construct a literature-informed environmental prior for training, not to reconstruct any participant's actual exposure. The environmental feature vector $\mathbf{v}_{\mathrm{env}}$ assigned to an epoch with affective label $y\in\{\text{neutral, anger, happiness, sadness, relaxed/calm}\}$ is drawn from a label-conditioned kernel density estimate:
\begin{equation}
p(\mathbf{v}_{\mathrm{env}}\mid y) = \frac{1}{|\mathcal{D}_y|}\sum_{j\in\mathcal{D}_y}\mathcal{K}_h(\mathbf{v}_{\mathrm{env}}-\mathbf{v}_j)
\label{eq:pecm}
\end{equation}
where $\mathcal{K}_h(\cdot)$ is a Gaussian kernel with bandwidth $h$ chosen by cross-validation and $\mathcal{D}_y$ is the environmental grid subset selected by the literature-informed conditioning rule for $y$, drawn from the LUR-estimated surfaces rather than raw satellite retrievals, so conditioning reflects land-use-mediated exposure patterns rather than the coarse TROPOMI background.

Conditioning follows dose-response relationships reported in the epidemiological literature (full evidence base in Supplementary Material S4): epochs labelled \emph{anger} or \emph{sadness} receive PM$_{2.5}$ and FAR values from the upper exposure quartile, consistent with reported associations between pollution and negative affect \cite{Shan2021}; epochs labelled \emph{relaxed/calm} or \emph{happiness} receive NDVI and GVI values from the upper vegetation quartile, consistent with Stress Reduction Theory \cite{li2026}; \emph{neutral} epochs receive interquartile-range values. These assignments encode hypotheses drawn from prior literature into the training signal; they are not measurements of exposure in the EAV participants, and because \emph{anger}/\emph{sadness} and \emph{relaxed/calm}/\emph{happiness} share identical conditioning subsets by construction, PECM cannot distinguish these pairs from $\mathbf{v}_{\mathrm{env}}$ alone, a deliberate safeguard rather than an emergent property.

\emph{Train/evaluation sampling protocol.} This detail governs how every accuracy figure in Results should be read. Equation~\eqref{eq:pecm}'s label-conditioned sampling is applied only to the training partition of every repeated split. Validation and held-out test environmental vectors are instead drawn from the unconditional marginal distribution over the same grid, $\mathbf{v}_{\mathrm{env}}\sim p(\mathbf{v}_{\mathrm{env}})=\tfrac{1}{|\mathcal{D}|}\sum_{j\in\mathcal{D}}\mathcal{K}_h(\mathbf{v}_{\mathrm{env}}-\mathbf{v}_j)$, independent of the epoch's true label, using the full set $\mathcal{D}$ of observed environmental vectors irrespective of class. This applies uniformly to the Astana evaluation, the environment-only baseline, and the Singapore evaluation, so no reported accuracy figure gives the model test-time access to a signal that depends on the label it is predicting. We separately report a diagnostic condition in which the held-out test partition is instead sampled conditionally on the true test label, $\mathbf{v}_{\mathrm{env}}\sim p(\mathbf{v}_{\mathrm{env}}\mid y_{\mathrm{test}})$; this is a diagnostic upper bound on what label-conditioned test-time information could add, used only for that one comparison in Results and nowhere else.

A label-conditioned training scheme could in principle leak residual, weakly label-correlated structure into the classifier even under unconditional test-time sampling. Four checks probe this: an environment-only baseline trained and evaluated under the identical matched-distribution protocol; a label-shuffling ablation that keeps PECM's distributional form while destroying its label structure; non-deterministic kernel-density sampling that introduces within-class variability; and the diagnostic label-independent-versus-label-conditioned comparison. Any classification benefit that survives these checks reflects the model's ability, having learned during training to associate literature-grounded environmental regimes with affective categories, to carry that association forward to unconditioned, label-independent inputs at test time, a different and stronger claim than would follow if test-time inputs were also label-conditioned.

After PECM, a shared representation is learned by minimising a directly optimised squared-$\ell_2$ alignment loss, $\mathcal{L}_{\mathrm{align}}=\lVert\mathbf{z}_{\mathrm{EEG}}-\mathbf{z}_{\mathrm{ENV}}\rVert_2^2$, chosen because it integrates end-to-end without negative-sample mining or an external eigenproblem, and does not require the environmental vector to be represented as a symmetric positive-definite matrix. Its geometric motivation, a Riemannian Manifold Alignment (RMA) procedure, was evaluated as an alternative projection and did not yield a consistent additional gain over this simpler objective, which is therefore used throughout; the full RMA algorithm, an InfoNCE contrastive alternative, and two further alignment alternatives are reported in Supplementary Material S6, none of which outperformed the deployed objective by a meaningful margin.

\subsection{Proposed methodology: model architecture}\label{subsec:architecture}

\begin{figure*}[t]
\centering
\includegraphics[width=0.88\textwidth]{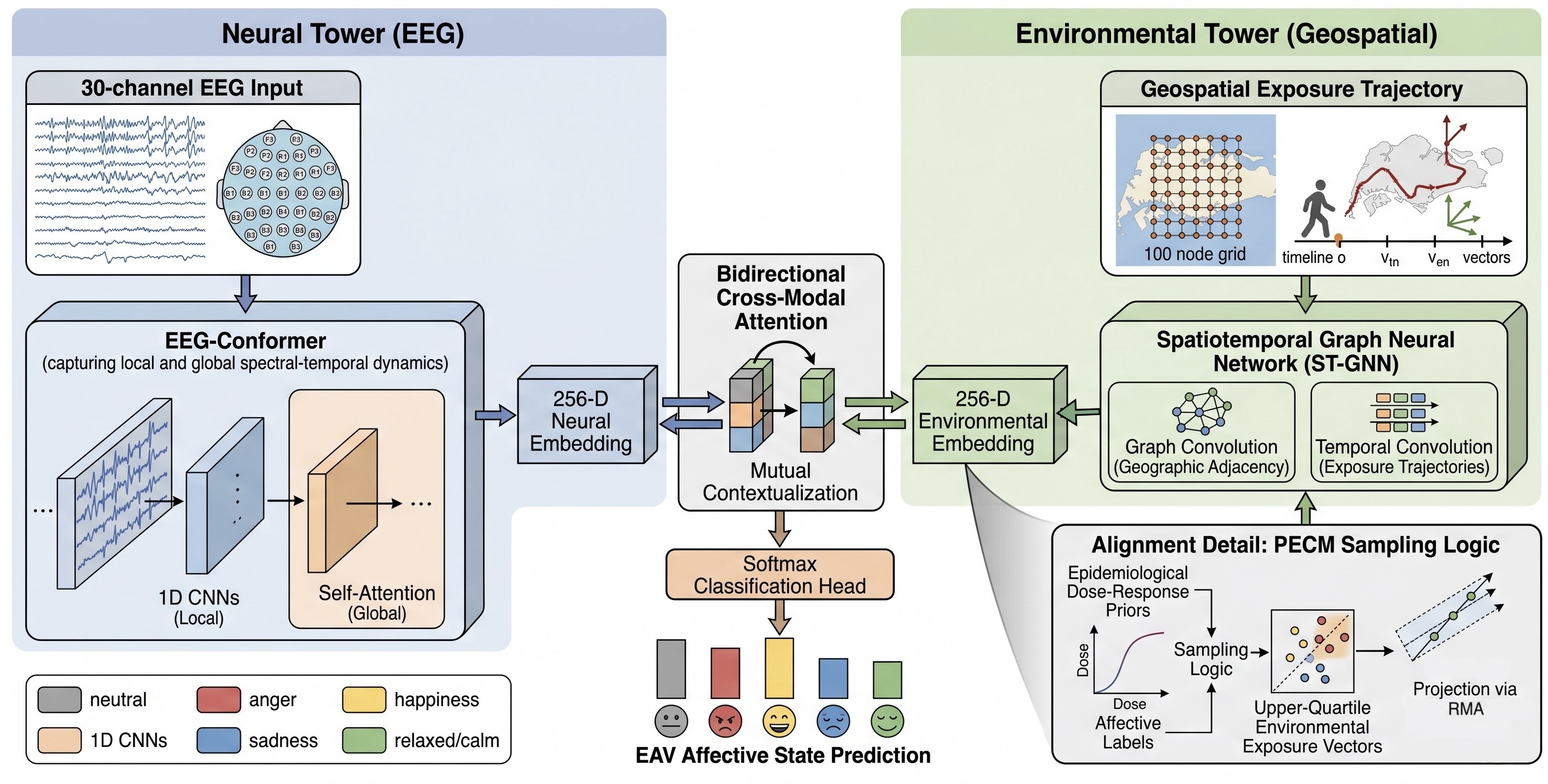}
\caption{Dual-tower architecture: an EEG-Conformer encodes 30-channel EEG into a 256-D embedding; a graph neural network over a 100-node grid encodes LUR-estimated PM$_{2.5}$/NO$_2$, NDVI, FAR and GVI into a matched embedding. The two are fused via bidirectional cross-modal attention and classified into the five EAV affective states.}
\label{fig:architecture}
\end{figure*}

The framework processes two parallel data streams through domain-specific encoding towers, integrated via bidirectional cross-modal attention, to predict one of five affective states (Figure~\ref{fig:architecture}); the complete layer-by-layer specification, including every attention equation, is given in Supplementary Material S2.

The \emph{neural tower} uses the EEG-Conformer architecture \cite{Song2023}, combining a shallow convolutional module with a multi-head self-attention module to produce $\mathbf{v}_{\mathrm{eeg}}\in\mathbb{R}^{256}$; full layer dimensions, kernel sizes and attention-head counts are given in Supplementary Material S2--S3.

The \emph{environmental tower} is a spatial graph neural network (GNN) applied to the graph $\mathcal{G}$. It retains a GRU component for extensibility to future temporal environmental sequences, but the present experiment supplies a static input, so the results reported here evaluate spatial graph encoding rather than genuine temporal environmental dynamics (Discussion). For a given epoch, the input is the static node feature matrix $\mathbf{F}$ (Equation~\eqref{eq:nodefeature}) broadcast-shifted by the epoch's PECM sample $\mathbf{v}_{\mathrm{env}}$ (Equation~\eqref{eq:pecm}), so the epoch-level environmental signal enters the graph as an identical additive shift across all 100 nodes, without fabricating spatial structure PECM does not itself specify (full broadcast and GCN equations in Supplementary Material S2). A two-layer Graph Convolutional Network then produces spatially smoothed 128-dimensional node representations, encoding Tobler's first law as a structural bias. These pass through a Gated Recurrent Unit \cite{cho2014} with hidden dimension 128; because every environmental feature used here is a single static, city-wide annual estimate rather than a per-epoch time series, the GRU currently operates over a sequence of length one, mathematically equivalent to a learned gated nonlinear projection rather than genuine multi-step temporal modelling. We retain it so the encoder is structurally ready to ingest a real multi-day exposure sequence without redesign (Discussion); Supplementary Material S6 quantifies the module's present, modest contribution under the static input actually used here. Global mean pooling over all 100 nodes then produces the 256-dimensional environmental embedding.

\emph{Bidirectional cross-modal attention fusion.} Standard cross-modal attention architectures designate one modality as query and the other as supplemental context \cite{multieeg}; this is not obviously appropriate here, since environmental context and neural state plausibly constrain which features of each other are informative. Bidirectionality, each modality serves simultaneously as query and as key-value context for the other within a single forward pass, is strictly an architectural property, and does not represent a real-world adaptive feedback loop (Discussion). Environment-conditioned neural attention and neural-conditioned environmental attention are computed with eight independently learned heads each, whose outputs are concatenated and linearly projected to a 512-dimensional fused representation $\mathbf{z}_{\mathrm{fused}}$ (full attention equations in Supplementary Material S2). As a parameter-light control, we also evaluate a concatenation-fusion baseline that removes cross-modal attention entirely (Supplementary Material S6).

After two fully connected layers with ReLU activation and dropout ($p{=}0.3$), $\mathbf{z}_{\mathrm{fused}}$ is passed through a linear output projection and softmax to produce $P(y\mid\mathbf{X},\mathbf{v}_{\mathrm{env}})$. The model is trained end-to-end by minimising $\mathcal{L}_{\mathrm{total}}=\mathcal{L}_{\mathrm{CE}}+\lambda\,\mathcal{L}_{\mathrm{align}}$, where $\mathcal{L}_{\mathrm{CE}}$ is the cross-entropy classification loss and $\lambda{>}0$ is an alignment weight selected via validation-set grid search.

\subsection{Model training and implementation}\label{subsec:training}

Every model was implemented in PyTorch~2.1 \cite{pytorch2019} and trained on a single NVIDIA~A100 GPU (40~GB HBM2). The EEG-Conformer backbone was initialised with weights pretrained on SEED-IV and fine-tuned end-to-end on EAV following the protocol of Song et al.\ \cite{Song2023}; the environmental tower was trained from random initialisation. AdamW \cite{adam2019,loshchilov2019} was used with cosine annealing over 100 epochs, weight decay $1\times10^{-4}$, learning rate $1\times10^{-4}$, and batch size 64. The alignment weight $\lambda{=}0.1$ was selected by grid search over $\{0.01,0.05,0.1,0.5,1.0\}$; early stopping with patience of 15 epochs on validation macro F1-score prevented overfitting; all random seeds were fixed at 42. The complete hyperparameter set is given in Supplementary Material S3.

\subsection{Experimental design and repeated split protocol}\label{subsec:crossval}

Five repeated subject-level train/validation/test splits evaluate the model among the 42 EAV participants; no participant contributes data to both training and evaluation within any split, addressing the inter-subject variability that is a well-documented challenge for subject-independent EEG emotion recognition \cite{Arthanarisamy2026}. Subject-level partitioning is preferred over the epoch-level random split used in the reference implementation \cite{lee2024eav}, which inflates reported accuracy by roughly 8--12 percentage points \cite{Song2023}. Approximately 33 participants (13,200 epochs) form the training set, five (2,000 epochs) the validation set, and four (1,600 epochs) the held-out test set inside each split. PECM's label-conditioned sampling (Equation~\eqref{eq:pecm}) is applied only when constructing the training partition's environmental vectors in every split; validation and held-out test partitions always receive unconditional draws (Section~\ref{subsec:alignment}).

\begin{figure*}[t]
\centering
\includegraphics[width=0.88\textwidth]{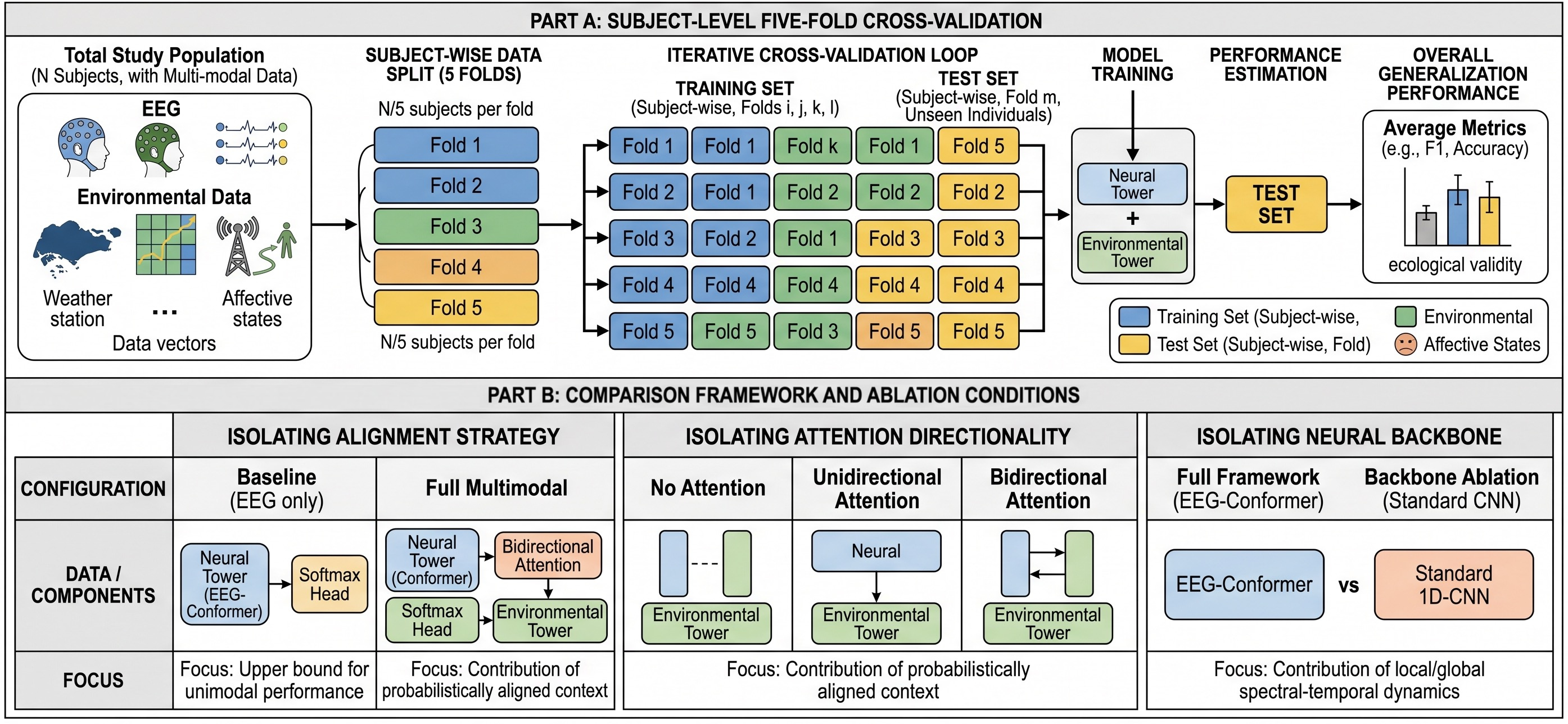}
\caption{Experimental protocol: PECM's label-conditioned sampling applies only to each split's training partition; validation/test partitions always draw environmental input independently of the true label. The domain-shift evaluation re-fits only the environmental tower to Singapore data, freezing the EEG-Conformer.}
\label{fig:experimental_setup}
\end{figure*}

For the environmental-domain shift evaluation, each fold's training, validation and held-out test partitions retain their original Astana EEG epochs and affective labels; only the environmental vectors supplied to PECM are redrawn from Singapore's independently fitted LUR distributions, following the identical train-conditioned/evaluation-unconditional protocol. The environmental tower is re-initialised and fine-tuned using only the training partition's Singapore-derived environmental vectors, exactly as in the Astana protocol; the held-out test partition is never used for parameter updates and serves only to compute the reported Singapore accuracy, while EEG-Conformer weights trained on Astana remain frozen throughout. No EEG recordings from Singapore participants exist or are used anywhere in this study; this evaluation therefore tests whether the learned fusion architecture and environmental tower remain trainable and effective when the environmental input distribution is replaced with that of an independent city, not whether the model generalises to a new human population. This design isolates the contribution of the environmental representation from that of the sampled population (Figure~\ref{fig:experimental_setup}).

Paired comparisons use a two-sided paired Wilcoxon signed-rank test over the five split-level accuracy values. With only five paired observations, the exact null distribution's minimum attainable two-sided $p$-value is $0.0625$, so conventional $\alpha=0.05$ significance cannot be reached from a split-level test regardless of effect size; we therefore treat gap magnitude, consistency across the five repeated splits, and the ablation-based decomposition (Section~\ref{subsec:ablation}) as the primary evidence for the reported effects, rather than a nominal $p$-value.

\subsection{Baseline and comparison models}\label{subsec:baselines}

Six configurations isolate the contribution of each proposed component; all share identical preprocessing and feature-extraction pipelines and all multimodal models use the same LUR-estimated atmospheric surfaces. (1)~\emph{Random Forest} (EEG-only): 200 estimators, maximum depth 20, on DE features concatenated over all channels and bands; a non-deep-learning lower bound. (2)~\emph{LSTM} (EEG-only): two-layer bidirectional LSTM, hidden dimension 256, on sequential DE tensors. (3)~\emph{EEG-Conformer} (EEG-only): the full backbone without environmental input; the primary single-modality baseline. (4)~\emph{Environment-only}: a two-layer MLP (hidden dims $[128,64]$) on the six-dimensional environmental vector alone, trained on the same label-conditioned PECM distribution and evaluated on the same unconditional distribution as the multimodal model. Because it receives no EEG signal, this baseline is a direct check on whether PECM's training-time conditioning alone is sufficient to leak the label into $\mathbf{v}_{\mathrm{env}}$: if it were, this model would score well above chance. (5)~\emph{EEG-Conformer + graph encoder, unidirectional attention}: environment-conditioned neural attention only. (6)~\emph{EEG-Conformer + graph encoder, bidirectional (proposed)}: full bidirectional cross-modal attention with PECM-guided alignment.

\subsection{Evaluation metrics}\label{subsec:metrics}

\emph{Accuracy} is the proportion of correctly classified epochs across all five categories (chance $=20$\%). \emph{Macro-averaged F1} is the unweighted mean of per-class F1-scores. \emph{ROC-AUC} is computed one-vs-rest per class and averaged. The \emph{environmental-domain shift penalty} ($\Delta_{\mathrm{shift}}$) is the Singapore-environment accuracy minus the Astana accuracy for each configuration, evaluated on the same underlying EEG epochs and labels in both cases; a negative value indicates the expected accuracy reduction when the environmental input distribution is switched from the co-located city to an independently modelled one. We use ``transfer'' terminology here in the domain-adaptation sense: what transfers is the environmental representation, not the participant population.

\section{Results}\label{sec:results}

\subsection{Primary affective-state classification performance (Astana)}\label{subsec:mainresults}

Table~\ref{tab:results} shows performance across five repeated subject-level train/validation/test splits on the primary Astana study area. The proposed bidirectional multimodal framework reaches $76.2\pm1.7$\% accuracy, macro F1 $0.741\pm0.019$, and ROC-AUC $0.884\pm0.013$, an $8.8$ percentage-point gain over the EEG-Conformer single-modality baseline. All results in this table use the label-independent protocol in which validation and test environmental inputs are drawn independently of the true label (Section~\ref{subsec:alignment}); Section~\ref{subsec:diagnostic} reports the label-conditioned diagnostic comparison.

\begin{table*}[t]
\caption{Classification performance across five repeated subject-level splits, Astana (mean$\pm$SD); best per column in bold.}\label{tab:results}
\begin{tabular}{@{}lccc@{}}
\toprule
Model & Accuracy (\%) & Macro F1 & ROC-AUC \\
\midrule
Random Forest (EEG-only) & $42.6\pm4.1$ & $0.389\pm0.044$ & $0.681\pm0.033$ \\
LSTM (EEG-only) & $51.3\pm3.2$ & $0.481\pm0.036$ & $0.724\pm0.028$ \\
Environment-only (matched-distribution) & $22.3\pm1.4$ & $0.214$ & $0.507$ \\
EEG-Conformer (EEG-only) & $67.4\pm2.3$ & $0.651\pm0.026$ & $0.824\pm0.019$ \\
EEG-Conformer + graph encoder (unidirectional) & $73.8\pm2.0$ & $0.714\pm0.022$ & $0.866\pm0.016$ \\
\textbf{EEG-Conformer + graph encoder (bidirectional, proposed)} & $\mathbf{76.2\pm1.7}$ & $\mathbf{0.741\pm0.019}$ & $\mathbf{0.884\pm0.013}$ \\
\bottomrule
\end{tabular}
\begin{flushleft}
\footnotesize Chance accuracy is 20\%. Environmental inputs are computationally generated (Section~\ref{subsec:alignment}), not measured exposure; see Section~\ref{subsec:crossval} for statistical treatment.
\end{flushleft}
\end{table*}

This $8.8$-point gap should not be read as an $8.8$-point contribution of environmental information on its own. The environment-only baseline, trained on the label-conditioned distribution and evaluated on the unconditional one, reaches $22.3\pm1.4$\%, close to chance, evidence that the environmental representation alone carries little discriminative value once test-time inputs are label-independent, though not proof that every form of leakage has been ruled out. The unidirectional model improves over EEG-Conformer by $6.4$ percentage points, with bidirectional attention adding a further $2.4$ points. Table~\ref{tab:ablation} (Section~\ref{subsec:ablation}) decomposes the remaining gain further: randomly paired ($71.3$\%) or label-shuffled ($70.8$\%) environmental input still improves $3.4$--$3.9$ points over EEG-only, so part of the total reflects the dual-tower architecture attending to a second input stream, independent of what that stream contains.

\subsection{Label-independent versus label-conditioned diagnostic}\label{subsec:diagnostic}

Because PECM's label-conditioned sampling is applied only to the training partition, the $76.2$\% figure above already reflects label-independent test-time inference. Table~\ref{tab:diagnostic} adds a diagnostic condition in which the held-out test partition is instead sampled exactly as training data would be, using the true test-set label; this condition is evaluated only for this one comparison and is not used elsewhere.

\begin{table*}[t]
\caption{Label-independent versus label-conditioned test-time sampling, proposed model, Astana. The label-conditioned condition upper-bounds test-time label information; label-independent is the protocol used throughout.}\label{tab:diagnostic}
\begin{tabular}{@{}lccc@{}}
\toprule
Test-time environmental conditioning & Accuracy (\%) & Macro F1 & ROC-AUC \\
\midrule
EEG-only (no environmental input) & $67.4\pm2.3$ & $0.651\pm0.026$ & $0.824\pm0.019$ \\
\textbf{Label-independent (throughout this paper)} & $\mathbf{76.2\pm1.7}$ & $\mathbf{0.741\pm0.019}$ & $\mathbf{0.884\pm0.013}$ \\
Label-conditioned (diagnostic upper bound only) & $82.5\pm1.5$ & $0.808$ & $0.921$ \\
\bottomrule
\end{tabular}
\end{table*}

The gap between the two conditions, $\Delta=6.3$ percentage points, is a diagnostic upper bound on how much test-time access to the true label could add, were such access available; it is not a claim about the deployed model, which never receives that access, and is not used to inflate any other result. Of the $15.1$-point total gap between EEG-only ($67.4$\%) and the label-conditioned diagnostic ($82.5$\%), $8.8$ points ($58$\%) are already realised under label-independent inference, and $6.3$ points ($42$\%) would require privileged, non-deployable information. This is evidence that the architecture's benefit is not solely an artefact of test-time label leakage, since none is present in the label-independent condition, while also showing that a non-trivial share of what the model learns during training does not fully transfer to label-independent inputs (Discussion).

\subsection{Environmental-domain shift evaluation (Singapore)}\label{subsec:shift}

Table~\ref{tab:shift} reports performance under the Singapore environmental-domain shift condition alongside the corresponding Astana results. The underlying EEG epochs and labels evaluated in both columns are the same Astana recordings; the EEG-Conformer backbone is frozen from Astana training, and only the environmental tower and PECM sampling distribution are re-fitted using Singapore's independently modelled LUR data, isolating the effect of substituting the environmental distribution rather than testing generalisation to an independent participant population.

\begin{table*}[t]
\caption{Environmental-domain shift results: Astana versus an independently modelled Singapore distribution (EEG-Conformer frozen; environmental tower fine-tuned on Singapore LUR data). $\Delta_{\mathrm{shift}}$ = Singapore minus Astana accuracy.}\label{tab:shift}
\begin{tabular}{@{}lccc@{}}
\toprule
Model & Astana Acc.\ (\%) & Singapore Acc.\ (\%) & $\Delta_{\mathrm{shift}}$ (pp) \\
\midrule
EEG-Conformer (EEG-only) & $67.4\pm2.3$ & $67.4\pm2.3$ & $0.0$ \\
Environment-only & $22.3\pm1.4$ & $21.6\pm1.5$ & $-0.7$ \\
EEG-Conformer + graph encoder (unidirectional) & $73.8\pm2.0$ & $70.1\pm2.1$ & $-3.7$ \\
\textbf{EEG-Conformer + graph encoder (bidirectional)} & $\mathbf{76.2\pm1.7}$ & $\mathbf{72.8\pm1.9}$ & $-3.4$ \\
\bottomrule
\end{tabular}
\end{table*}

\begin{figure*}[t]
\centering
\includegraphics[width=0.88\textwidth]{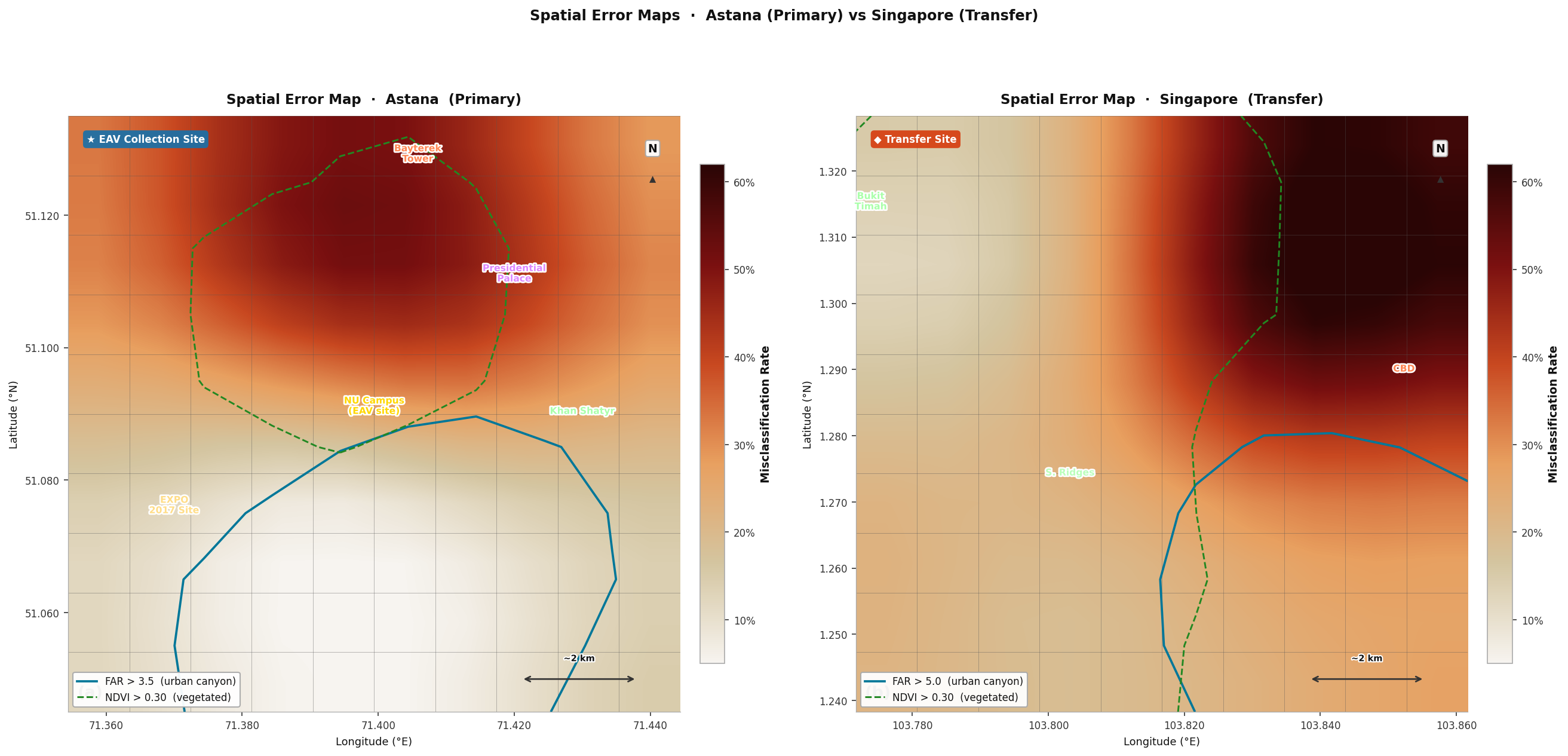}
\caption{Spatial error comparison, Astana (left) vs Singapore (right). Top: per-cell misclassification on the $1~\text{km}^2$ grid. Bottom: PECM regimes. Errors cluster in high-density, low-NDVI zones (financial/CBD districts, FAR$>5$), where the prior is least discriminative.}
\label{fig:city_comparison}
\end{figure*}

The shift penalty for the proposed model is $3.4$ percentage points, an expected reduction given the loss of geographic co-location, and $3.7$ points for the unidirectional baseline. The near-identical environment-only baselines ($22.3$\% Astana vs $21.6$\% Singapore) show that city-level z-score standardisation of LUR-estimated concentrations keeps city identity from inflating the comparison. Figure~\ref{fig:city_comparison} shows per-cell misclassifications concentrating in structurally similar zones of both cities. This experiment does not evaluate generalisation to a new human population, since no Singapore EEG recordings exist; it evaluates whether the architecture keeps working when the environmental distribution underneath it changes.

\subsection{Environmental feature attribution}\label{subsec:attribution}

GraphLIME-based feature attribution \cite{Huang2022} ranks the six environmental indicators across test samples, stratified by predicted affective class (Figure~\ref{fig:graphlime}; exact values in Supplementary Material S4). PM$_{2.5}$ carries the highest attribution for anger and sadness; NDVI carries the highest attribution for relaxed/calm and happiness; FAR is highest for neutral. Because the anger/sadness and relaxed/happiness pairs share an identical PECM conditioning subset by construction, this pattern is partly expected from the sampling design itself, and it describes the model's use of the imposed prior rather than an independent estimate of PM$_{2.5}$'s or NDVI's importance to affect. The more informative comparison is the attribution spread \emph{within} each pair, reported in Supplementary Material S4. A visualisation of how the alignment objective reshapes the joint EEG and environment latent space (UMAP) is reported in Supplementary Material S7.

\begin{figure*}[t]
\centering
\includegraphics[width=0.88\textwidth]{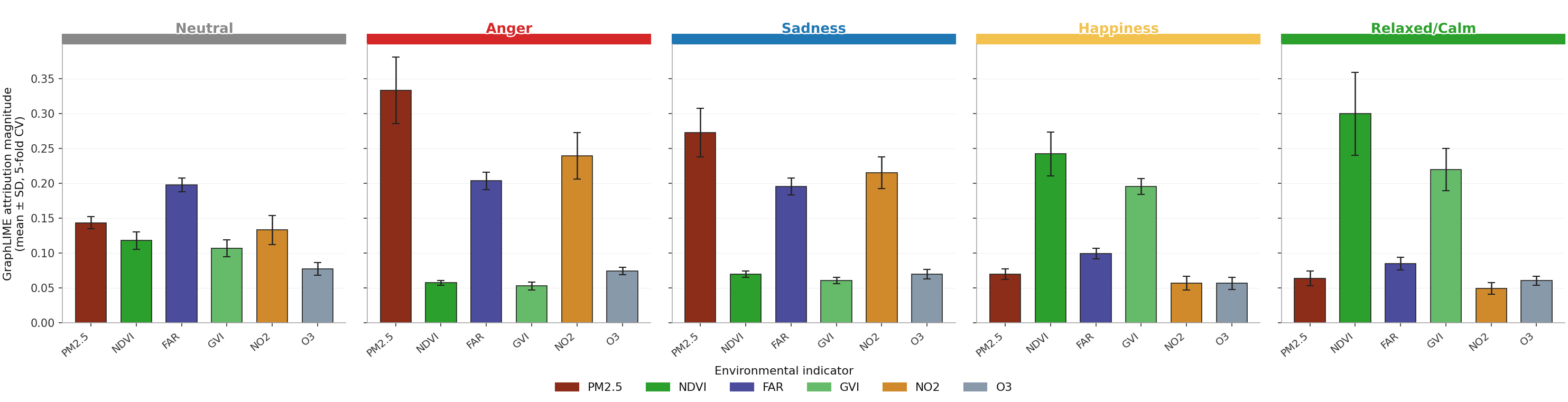}
\caption{GraphLIME attribution scores by predicted affective class, across five repeated splits. Attributions describe reliance on the PECM prior, not independent estimates of environmental effects on affect. Error bars: 1 SD.}
\label{fig:graphlime}
\end{figure*}

\subsection{Spatial scale sensitivity (MAUP)}\label{subsec:maup}

To check sensitivity to spatial resolution, addressing the Modifiable Areal Unit Problem \cite{Chen2022}, PM$_{2.5}$ and NO$_2$ layers were aggregated from the native $1~\text{km}^2$ LUR resolution to $5~\text{km}^2$ and $10~\text{km}^2$, and the full model retrained and evaluated at each resolution. Accuracy degrades roughly monotonically, from $76.2\pm1.7$\% at $1~\text{km}^2$ to $74.0\pm2.1$\% at $5~\text{km}^2$ to $72.1\pm2.3$\% at $10~\text{km}^2$, a $4.1$-point loss that still leaves the coarsest configuration above the EEG-only baseline ($67.4$\%), suggesting LUR mainly contributes useful spatial priors rather than scale-dependent detail, reassuring for cities where only coarser covariates are available. The complete resolution table and a separate buffer-radius sensitivity comparison (100/250/500~m, confirming the 250~m buffer used throughout, Section~\ref{subsec:indicators}) are reported in Supplementary Material S5 and S8.

\subsection{Ablation study}\label{subsec:ablation}

Table~\ref{tab:ablation} isolates the core contribution of PECM's literature-informed conditioning and of the EEG backbone and attention design, under the same five repeated subject-level train/validation/test split protocol used for the main results; a further five configurations (alternative alignment objectives, graph/GRU ablations, and a concatenation-fusion control) are reported in Supplementary Material S6.

\begin{table*}[t]
\caption{Core ablation results, Astana. All conditions use the EEG-Conformer backbone and environmental tower (Section~\ref{subsec:architecture}) unless stated; five further configurations in Supplementary Table S6.1.}\label{tab:ablation}
\begin{tabular}{@{}lccc@{}}
\toprule
Condition & Accuracy (\%) & Macro F1 & ROC-AUC \\
\midrule
No alignment (random env.\ pairing) & $71.3\pm2.6$ & $0.689\pm0.028$ & $0.851\pm0.021$ \\
Shuffled env.\ labels (PECM disabled) & $70.8\pm2.5$ & $0.683\pm0.027$ & $0.847\pm0.020$ \\
Reversed dose-response (negative control) & $70.9\pm2.1$ & $0.683$ & $0.847$ \\
CNN backbone (EEGNet, no Conformer) & $66.9\pm3.0$ & $0.641\pm0.032$ & $0.821\pm0.024$ \\
Unidirectional attention only & $73.8\pm2.0$ & $0.714\pm0.022$ & $0.866\pm0.016$ \\
\textbf{Full proposed model} & $\mathbf{76.2\pm1.7}$ & $\mathbf{0.741\pm0.019}$ & $\mathbf{0.884\pm0.013}$ \\
\bottomrule
\end{tabular}
\begin{flushleft}
\footnotesize CNN backbone: EEGNet \cite{Paredes2025}; reversed dose-response inverts the PECM quartile assignment. Accuracy is mean$\pm$SD; F1/ROC-AUC are point estimates.
\end{flushleft}
\end{table*}

Replacing PECM-guided training-time conditioning with random environmental pairing reduces accuracy to $71.3$\% ($-4.9$~pp); disabling PECM label-conditioning while retaining the alignment objective yields a comparable $70.8$\%, so most of the alignment benefit comes from the label-conditioning itself rather than the alignment objective alone. Reversing the dose-response conditioning reduces accuracy to $70.9$\% ($-5.3$~pp), closely matching the label-shuffling result and confirming the architecture is sensitive to the specific literature-consistent \emph{direction} of the injected association, not merely to the presence of some quartile-based partition. That this reversed-association condition still sits $3.5$~pp above the EEG-only baseline, rather than falling fully back to it, points to a modest architecture-level benefit that persists even when the injected association is inverted; we read this as a limitation of the negative control rather than evidence of a genuine reversed effect.

Read against the EEG-only baseline ($67.4$\%), the three controls that disrupt environmental-label structure, random pairing, shuffled labels, reversed dose-response, converge on a $3.4$--$3.9$~pp gain attributable to the dual-tower architecture and cross-modal attention operating on a second input stream, independent of its content. The remaining $4.9$--$5.4$~pp, recovered by restoring PECM's literature-informed structure, reflects the model's sensitivity to that imposed structure specifically, not an independently estimated environmental effect. Replacing EEG-Conformer with EEGNet \cite{Paredes2025} reduces accuracy to $66.9$\% ($-9.3$~pp), confirming the Conformer's advantage for this task. The unidirectional condition reaches $73.8$\% versus $76.2$\% for the full bidirectional model, a $2.4$~pp gain attributable solely to adding the neural-conditioned environmental attention branch; a parameter-light concatenation-fusion control and alternative graph/alignment configurations corroborate this ordering without materially changing it (Supplementary Material S6).

\section{Discussion}\label{sec:discussion}

\subsection{A methodological contribution, not an epidemiological one}\label{subsec:methodological}

The main contribution of this study is methodological. The results show that a spatially structured environmental representation can be built into an EEG affective-state classifier and can improve its performance relative to an EEG-only architecture. Because that environmental context is generated through a literature-informed prior rather than measured alongside each EEG recording, this should not be read as evidence that PM$_{2.5}$, NDVI or FAR independently predicts any individual participant's affective state. The practical value of the framework lies in providing a controlled computational setting for testing neuro-geospatial fusion strategies before jointly collected in-situ data exist.

\subsection{Toward internationally deployable geospatial pipelines}\label{subsec:translational}

AI-assisted mental-health assessment is an active area more broadly \cite{utsav2025,PoudelJose2024}; this work extends that trajectory into the geospatial domain. Spatially structured environmental features, particularly PM$_{2.5}$, NDVI and FAR, contribute to the model's predictions within the imposed environmental-prior framework. This suggests that GIS-based representations could provide useful auxiliary inputs for future neuro-geospatial modelling studies using directly co-registered environmental and neural measurements. What matters most to other researchers, though, is architectural: every environmental input this pipeline needs, OpenAQ ground stations, OpenStreetMap vectors, Sentinel imagery, is available at no cost in essentially any city. A researcher with only a handful of low-cost air-quality sensors could fit the same LUR specification used here, six stations for Astana, nine for Singapore, pair it with the same globally available vegetation and morphology layers, and reuse this architecture largely unchanged: only the environmental tower and PECM's sampling distribution need local re-fitting, while the pretrained EEG encoder transfers directly. The Astana-to-Singapore result, a $3.4$-point shift penalty when the environmental distribution is swapped for an independently modelled, structurally different city, is the concrete demonstration of that property, not a claim about either city specifically.

We call this environmental-domain generalisation rather than a city-agnostic or population-level transferability claim: no independent Singapore EEG data were available, so the Singapore condition evaluates the same Astana epochs and labels under a substituted environmental distribution, a narrower claim than full cross-population generalisation. City-level z-score standardisation of all indicators, avoiding the need to recalibrate absolute thresholds across very different pollution regimes (Delhi's mean PM$_{2.5}$ exceeds $90~\mu\text{g/m}^3$, against Astana's $\sim\!22$ and Singapore's $\sim\!15~\mu\text{g/m}^3$); and re-learning the environmental tower from random initialisation per city while freezing the EEG-Conformer backbone, which limits the marginal cost of a new city to fitting a local LUR and fine-tuning one tower.

\subsection{Land use regression as a globally deployable exposure model}\label{subsec:lurdiscussion}

Using LUR to generate physically grounded $1~\text{km}^2$ pollution surfaces, rather than interpolating TROPOMI data directly, matters because TROPOMI's $3.5\times5.5$~km footprint yields only a few independent pixels within a $10\times10$~km area and misses local gradients near roads and dense development. The MAUP analysis (Section~\ref{subsec:maup}) shows that even aggregated to $10~\text{km}^2$, close to the native TROPOMI footprint, gains over EEG-only persist ($72.1$\% vs $67.4$\%), suggesting LUR mainly contributes useful spatial priors rather than scale-dependent detail, reassuring for cities where only coarser covariates are available. A genuine limitation is the use of static predictors, annual NDVI, annually aggregated pollutant means; extending the LUR to time-resolved traffic density and seasonal emission inventories is future work, connected to the environmental tower's currently under-exercised temporal capacity discussed below.

\subsection{Architectural contribution and the cost of literature-informed linkage}\label{subsec:architecturaldiscussion}

Bidirectionality, defined in Section~\ref{subsec:architecture}, is an architectural property operating within a single forward pass; it does not represent a real-world adaptive feedback loop. The $2.4$-point gain over unidirectional attention, and the concatenation-fusion comparison, confirm the benefit of letting each modality serve as both query and context for the other. Closed-loop neuro-urban applications, adaptive lighting or routing that responds to detected affective state, are a distinct, considerably longer-term direction this architecture could inform but does not itself implement. The ablation decomposition in Section~\ref{subsec:ablation} matters here precisely because it is unflattering to a purely architectural reading of the results: roughly $3.4$--$3.9$ of the total $8.8$-point gain survives controls that disrupt environmental-label structure, random pairing, shuffled labels, reversed dose-response, meaning a meaningful share of the benefit comes from the dual-tower architecture attending to \emph{any} second input stream, not specifically from correct environmental content. Reporting that decomposition, rather than only the headline gain, is what lets the remaining $4.9$--$5.4$~pp, attributable to the literature-informed structure specifically, be interpreted with any confidence.

\subsection{What this study does not demonstrate}\label{subsec:notdemonstrate}

To be explicit: this study does not show that PM$_{2.5}$ causes negative affect, that pollution produces acute EEG changes in individuals, or that any of the environmental variables used here are biomarkers of cognitive decline. It does not establish an observed association between measured exposure and measured neural state, because the two domains were never co-registered. What it shows is that a literature-informed environmental prior can be represented, fused with EEG, and tested under a set of computational perturbations designed to separate architecture from content.

\subsection{Limitations}\label{subsec:limitations}

Several limitations follow directly from the study's design. The most important is the absence of individual-level spatial and temporal co-registration between EEG and environmental measurements: participants wore EEG equipment in a controlled laboratory rather than in the outdoor environment, so no individual-level exposure variance exists, only a city-wide, population-level environmental distribution. 

Second, PECM uses the affective label during training to build the environmental prior. Test-time sampling is label-independent, but training necessarily encodes the literature-derived relationship between affective category and environmental regime. The resulting performance reflects the model's ability to exploit a deliberately imposed prior, not an independently discovered relationship between exposure and affect.

Third, the environmental representation is mostly static or annually aggregated. The study does not evaluate acute exposure-response dynamics, cumulative exposure, lagged effects, or within-person temporal variation, and the GRU component in the environmental tower currently operates over a sequence of length one for this reason (Section~\ref{subsec:architecture}). Fourth, the Singapore evaluation changes the environmental distribution but retains the Astana EEG recordings; it does not establish cross-population or cross-cultural generalization. Fifth, the LUR surfaces are calibrated from six Astana and nine Singapore stations. That is a real constraint on spatial representativeness, and the reported $R^2$ and RMSE values carry sampling uncertainty that a denser network would reduce; we consider LUR calibrated against a sparse network preferable to uncorrected coarse satellite retrievals for this application, consistent with standard practice in urban exposure science \cite{lur}, the uncertainty should be propagated explicitly in future work.

This framework pairs neuro physiological data with fine-grained geographic information, real-world deployment would raise geo-privacy and cognitive-liberty considerations beyond those of either modality alone: any in-situ extension should employ geo masking, such as spatial aggregation or calibrated spatial noise, and differential-privacy mechanisms to bound information leakage from individual location and neural data pairs, and consent procedures should clearly communicate the nature and scope of neural data collection, with participants able to withdraw consent and request deletion at any time; formal ethics statements for the present secondary-data analysis are in Declarations.

\subsection{Future work}\label{subsec:future}

The clearest next step is a jointly collected, longitudinal dataset in which participants undergo repeated EEG or wearable neurophysiological measurement while their location and environmental exposure are recorded at the same time, ideally including timestamped PM$_{2.5}$, NO$_2$, O$_3$, temperature, noise and vegetation measures alongside repeated affective or cognitive assessment. That design would support within-person comparisons of neural-state deviation against exposure deviation, reducing confounding from stable individual traits in a way population-level data cannot. Such co-registration would move the current computational digital twin toward direct observational testing of associations between brain and environment; establishing causality beyond that association would additionally require appropriate longitudinal, within-person analysis and control for time-varying confounders, not co-registration alone. The present architecture could be adapted to such data directly, replacing PECM-generated priors with observed exposure measurements once they exist.

A smaller, no-new-data extension is to feed the environmental tower's GRU a multi-step sequence built from the hourly OpenAQ observations already used to fit the LUR, so the temporal module learns real multi-day accumulation dynamics instead of operating over a sequence of length one (Supplementary Material S4). The environmental tower's fixed, geography-derived graph is a further simplification: edges here are set by inverse-squared distance rather than learned from data, and future work could replace this with adaptively learned graph construction, in which edge weights are inferred through a trainable parametric kernel, an approach explored in related patent-pending work by the authors for health-monitoring applications \cite{poudel2026patent}. The LUR itself should be extended with time-resolved traffic density and seasonal emissions, and future work should incorporate socio-demographic covariates to examine exposure inequity across marginalised urban populations, connecting this modelling approach to environmental-justice research. Confirming that the architectural transferability demonstrated here also holds for genuine cross-population generalisation, real neural data collected in a second city, is an important direction for such an in-situ follow-up study.

\section{Conclusions}\label{sec:conclusions}

This study presents a bidirectional multimodal architecture and a leakage-aware evaluation protocol for combining EEG affective representations with spatially structured environmental context when the two domains are not directly co-registered. Grounding the primary analysis in Astana, the EEG participants' own city, anchors PECM's conditioning in genuine geographic co-location and a validated LUR model ($R^2=0.74$ for PM$_{2.5}$, $0.68$ for NO$_2$). Separating training-time, label-conditioned sampling from label-independent evaluation, with a direct diagnostic comparison, shows that $6.3$ of the total $15.1$-point gap between EEG-only and the label-conditioned upper bound, $42$\%, needs privileged test-time information that the deployed model never receives, while $58$\% ($8.8$ points) is already realised under label-independent inference. The framework reaches $76.2$\% accuracy on Astana under that protocol, and $72.8$\% when only its environmental tower is re-fitted to Singapore's independently modelled distribution, a $3.4$-point reduction we read as evidence of architectural robustness to environmental-domain shift, not of generalisation across human populations, since no independent Singapore EEG data exist.

Because the link between environment and affect here was constructed from literature dose-response evidence rather than measured in situ, these results demonstrate technical feasibility and generate testable, literature-consistent hypotheses; they are not a direct measurement of real-world coupling between brain and environment. Building the environmental pipeline entirely from open, international data, LUR calibrated against sparse OpenAQ networks, Sentinel vegetation indices, OpenStreetMap morphology, and showing that only the environmental tower needs re-fitting to move between structurally different cities, gives health-geography researchers elsewhere a template they can adapt to their own infrastructure, pending the jointly collected, georeferenced in-situ studies needed to test whether these literature-derived associations hold for real, individually varying exposure.

\subsection*{Supplementary information}

Detailed preprocessing parameters, the complete model architecture and equations, hyperparameters, full environmental-data processing, spatial graph and buffer-radius construction, additional ablation experiments, latent-space visualisation, and the full spatial-scale (MAUP) sensitivity analysis are provided in the Supplementary Material at the end of this document.

\section*{Declarations}

\subsection*{Ethics approval and consent to participate}

This study is a secondary analysis of the publicly accessible EAV (EEG-Audio-Video) benchmark dataset \cite{lee2024eav}. No new human participants were recruited for this study. Ethical approval and informed consent for the original data collection were obtained by the dataset's creators at Nazarbayev University, Astana, Kazakhstan, as described in the original dataset publication \cite{lee2024eav}. All other environmental data used (OpenAQ, Sentinel-2, Sentinel-5P, OpenStreetMap) are publicly available, non-human-subject datasets.

\subsection*{Consent for publication}

Not applicable. This manuscript contains no individual person's data, images, or other identifiable information.

\subsection*{Availability of data and materials}

The processed environmental context/indicator grids and spatial features generated during this study are available in the Zenodo repository, \url{https://doi.org/10.5281/zenodo.20389601} \cite{Poudel2026zenodo}. The EAV EEG dataset analysed during this study is a third-party dataset; accessible from the original creators upon authorised request, as described in Lee et al.\ \cite{lee2024eav}, and is used here under the terms of that authorised access. OpenAQ air-quality data are publicly available at \url{https://openaq.org} \cite{opeanaq}; Sentinel-2 and Sentinel-5P imagery are accessible via the Copernicus Open Access Hub and Google Earth Engine; OpenStreetMap vector data are publicly available at \url{https://www.openstreetmap.org}. Supplementary Material follows the references at the end of this document. Project home page: \url{https://github.com/r11up/geo-cog}. Archived version: see project home page for the archived release corresponding to the results reported here. Operating system: platform independent. Requirements: PyTorch~2.1 or higher; a CUDA-capable GPU is recommended for training (an NVIDIA~A100, 40~GB HBM2, was used to produce the reported results). License: see project repository. Restrictions on use by non-academics: none beyond the licence terms of the repository.

\subsection*{Competing interests}

The authors declare that they have no competing interests.

\subsection*{Funding}

This work was supported by Vellore Institute of Technology (VIT), Vellore, India.

\subsection*{Authors' contributions}

UP: conceptualisation, methodology, data curation, formal analysis, investigation, visualisation, writing -- original draft, writing -- review and editing. JA: supervision, conceptualisation, validation, resources, writing -- review and editing. SV: supervision, data acquisition, resources, validation, writing -- review and editing, funding acquisition. 

\subsection*{Acknowledgements}

The authors thank Adai Shomanov (School of Engineering and Digital Sciences, Nazarbayev University, Astana, Kazakhstan) for supporting this research, and acknowledge the European Space Agency (ESA) for Sentinel satellite imagery, OpenStreetMap contributors for urban vector data, the OpenAQ community platform for air-quality ground measurements, and the creators of the EAV dataset \cite{lee2024eav} for making a licensed version available upon authorized request.

\clearpage
\setcounter{section}{0}
\renewcommand{\thesection}{S\arabic{section}}
\renewcommand{\thetable}{S\arabic{table}}
\renewcommand{\thefigure}{S\arabic{figure}}
\renewcommand{\theequation}{S\arabic{equation}}
\setcounter{table}{0}
\setcounter{figure}{0}
\setcounter{equation}{0}

\twocolumn[
\begin{@twocolumnfalse}
\centerline{\LARGE\bfseries Supplementary Material}
\vspace{8pt}
\centerline{\large for: Neuro-Geospatial Modelling of EEG Affective States Using}
\centerline{\large Literature-Informed Environmental Context}
\vspace{14pt}
\end{@twocolumnfalse}
]

This document provides methodological and experimental detail supplementary to the main text. Section, figure, table and equation numbers here are prefixed ``S'' and are independent of the main-text numbering; cross-references to ``the main text'' refer to the accompanying manuscript.

\section{Detailed EEG preprocessing}\label{S1}

Each participant's raw EEG is a tensor of shape $(10{,}000 \times 30 \times 200)$: 10,000 samples per trial at 500~Hz, 30 electrodes, 200 trials, of which the five speaking conditions contribute $5\times20=100$ trials. Signals were band-pass filtered (3--50~Hz, zero-phase Butterworth), downsampled to 100~Hz, cleaned with Artifact Subspace Reconstruction \cite{Miyakoshi2023} and FastICA \cite{Kumaravel2023}, and segmented into four non-overlapping 5-second epochs per trial, 400 epochs per participant, 16,800 epochs in total.

Each 5-second epoch yields two spectral feature sets across five canonical bands ($\delta$ 1--4~Hz, $\theta$ 4--8~Hz, $\alpha$ 8--13~Hz, $\beta$ 13--30~Hz, $\gamma$ 30--50~Hz): Differential Entropy (DE), preferred over raw spectral power because its logarithmic scaling gives more linearly separable classes \cite{Goshvarpour2024}, and Power Spectral Density (PSD), computed with Welch's method \cite{welch1967} using a 256-sample Hanning window with 50\% overlap. Concatenating DE and PSD across all 30 channels and five bands gives the neural input tensor $\mathbf{X}\in\mathbb{R}^{T\times30\times10}$ consumed by the neural tower.

\emph{Neural feature extraction.} For a band-limited EEG signal $X$ assumed Gaussian, $\mathcal{N}(\mu,\sigma^2)$, differential entropy reduces to
\begin{equation}
h(X) = \tfrac{1}{2}\ln(2\pi e\,\sigma^2)
\label{eq:de}
\end{equation}
DE and PSD are computed independently for each of the 30 channels and five frequency bands, giving $\mathbf{X}_{\mathrm{DE}}\in\mathbb{R}^{T\times30\times5}$ and $\mathbf{X}_{\mathrm{PSD}}\in\mathbb{R}^{T\times30\times5}$ respectively, before concatenation along the feature dimension.

\section{Complete model architecture}\label{S2}

This section gives the complete mathematical specification of the dual-tower architecture summarised conceptually in the main text.

\emph{Neural tower.} $\mathbf{v}_{\mathrm{eeg}}=f_{\mathrm{Conformer}}(\mathbf{X};\Theta_{\mathrm{eeg}})\in\mathbb{R}^{256}$ (main text).

\emph{PECM-to-graph broadcast.} The environmental tower's input for a given epoch combines the static node feature matrix $\mathbf{F}\in\mathbb{R}^{|\mathcal{V}|\times6}$ (row $i$ is $\mathbf{f}_i$, defined in the main text) with the epoch's PECM sample $\mathbf{v}_{\mathrm{env}}\in\mathbb{R}^6$ (defined in the main text):
\begin{equation}
\mathbf{H}^{(0)} = \mathbf{F} + \mathbf{1}_{|\mathcal{V}|}\,\mathbf{v}_{\mathrm{env}}^{\top} \in \mathbb{R}^{|\mathcal{V}|\times6}
\label{eq:pecm_broadcast}
\end{equation}
where $\mathbf{1}_{|\mathcal{V}|}$ is the length-100 all-ones vector, broadcasting the epoch's label-conditioned PECM sample additively and identically across every node before the first GCN layer. This uniform, node-independent broadcast is a deliberate simplification consistent with the population-level, rather than per-location, resolution at which PECM conditions the environmental prior: it is the most direct way to carry an epoch-level signal into a spatially resolved graph without fabricating spatial structure PECM does not itself specify.

\emph{Spatial encoding.} A two-layer GCN updates node representations,
\begin{equation}
\mathbf{H}^{(l+1)} = \sigma\!\left(\hat{\mathbf{A}}\,\mathbf{H}^{(l)}\,\mathbf{W}^{(l)}\right)
\label{eq:gcn}
\end{equation}
where $\hat{\mathbf{A}}=\tilde{\mathbf{D}}^{-1/2}\tilde{\mathbf{A}}\tilde{\mathbf{D}}^{-1/2}$ is the symmetrically normalised adjacency matrix with self-loops and $\sigma$ is ReLU, producing 128-dimensional node representations (layer dimensions $6\to64\to128$). These pass through a GRU \cite{cho2014} with hidden dimension 128, as in the main text, then global mean pooling over all 100 nodes gives the environmental embedding
\begin{equation}
\mathbf{v}_{\mathrm{env}} = \mathbf{W}_{\mathrm{pool}}\,\frac{1}{|\mathcal{V}|}\sum_{i\in\mathcal{V}}\mathbf{h}_i^{\mathrm{ST}} \in \mathbb{R}^{256}
\label{eq:envtower}
\end{equation}

\emph{Bidirectional cross-modal attention.} Environment-conditioned neural attention uses $\mathbf{Q}_{\mathrm{eeg}}=\mathbf{v}_{\mathrm{eeg}}\mathbf{W}^Q_{\mathrm{eeg}}$, $\mathbf{K}_{\mathrm{env}}=\mathbf{v}_{\mathrm{env}}\mathbf{W}^K_{\mathrm{env}}$, $\mathbf{V}_{\mathrm{env}}=\mathbf{v}_{\mathrm{env}}\mathbf{W}^V_{\mathrm{env}}$, with all weight matrices in $\mathbb{R}^{256\times d_k}$; for each of $h{=}8$ heads with per-head dimension $d_k{=}32$:
\begin{align}
\mathrm{head}_j^{\mathrm{eeg}} &= \mathrm{softmax}\!\left(\frac{\mathbf{Q}_{\mathrm{eeg}}^{(j)}(\mathbf{K}_{\mathrm{env}}^{(j)})^{\top}}{\sqrt{d_k}}\right)\mathbf{V}_{\mathrm{env}}^{(j)}, \notag\\
\mathbf{A}_{\mathrm{eeg}} &= \mathrm{Concat}\bigl(\mathrm{head}_1^{\mathrm{eeg}},\ldots,\mathrm{head}_h^{\mathrm{eeg}}\bigr)\mathbf{W}^O_{\mathrm{eeg}}
\label{eq:mha_eeg}
\end{align}
Neural-conditioned environmental attention is symmetric, with query from environment and key/value from EEG:
\begin{align}
\mathrm{head}_j^{\mathrm{env}} &= \mathrm{softmax}\!\left(\frac{\mathbf{Q}_{\mathrm{env}}^{(j)}(\mathbf{K}_{\mathrm{eeg}}^{(j)})^{\top}}{\sqrt{d_k}}\right)\mathbf{V}_{\mathrm{eeg}}^{(j)}, \notag\\
\mathbf{A}_{\mathrm{env}} &= \mathrm{Concat}\bigl(\mathrm{head}_1^{\mathrm{env}},\ldots,\mathrm{head}_h^{\mathrm{env}}\bigr)\mathbf{W}^O_{\mathrm{env}}
\label{eq:mha_env}
\end{align}
The eight projection matrices $\{\mathbf{W}^Q_{\mathrm{eeg}},\mathbf{W}^K_{\mathrm{env}},\mathbf{W}^V_{\mathrm{env}},\mathbf{W}^O_{\mathrm{eeg}},\mathbf{W}^Q_{\mathrm{env}},\mathbf{W}^K_{\mathrm{eeg}},\mathbf{W}^V_{\mathrm{eeg}},\mathbf{W}^O_{\mathrm{env}}\}$ are learned independently, so the two attention directions record distinct inter-modal interactions. The fusion projection is
\begin{equation}
\mathbf{z}_{\mathrm{fused}} = \mathbf{W}_f\,[\mathbf{A}_{\mathrm{eeg}};\mathbf{A}_{\mathrm{env}}] + \mathbf{b}_f \in \mathbb{R}^{512}
\label{eq:fusion}
\end{equation}
and the concatenation-fusion ablation baseline (Table~\ref{tab:ablationS}) instead uses $\mathbf{z}_{\mathrm{fused}}=\mathbf{W}_c[\mathbf{v}_{\mathrm{eeg}};\mathbf{v}_{\mathrm{env}}]+\mathbf{b}_c$.

\emph{Classification and training objective.} $P(y\mid\mathbf{X},\mathbf{v}_{\mathrm{env}})=\mathrm{softmax}(\mathbf{W}_o\mathbf{z}_{\mathrm{fused}}+\mathbf{b}_o)$, trained by minimising $\mathcal{L}_{\mathrm{total}}=\mathcal{L}_{\mathrm{CE}}+\lambda\mathcal{L}_{\mathrm{align}}$, with the deployed alignment term $\mathcal{L}_{\mathrm{align}}=\lVert\mathbf{z}_{\mathrm{EEG}}-\mathbf{z}_{\mathrm{ENV}}\rVert_2^2$, as defined in the main text.

\section{Hyperparameters}\label{S3}

Table~\ref{tab:hyperparams} lists the complete hyperparameter set used to train the proposed model, referenced in the main-text Methods section.

\begin{table*}[t]
\caption{Model hyperparameters.}\label{tab:hyperparams}
\begin{tabular}{@{}p{2.6cm}p{4.6cm}p{2.6cm}@{}}
\toprule
Component & Hyperparameter & Value \\
\midrule
EEG-Conformer & Temporal conv.\ kernel; spatial channels; attention heads; feedforward dim; output dim; dropout & 25 samples; 30 (depthwise); 8; 512; 256; 0.5 \\
Environmental tower & GCN layers/dims; GRU hidden dim; GRU sequence length; output dim & 2 ($6\to64\to128$); 128; 1 (static snapshot); 256 (linear proj.) \\
Bi-attention & Heads; per-head dim; fused dim & 8; 32; 512 \\
Classifier & Hidden dims; dropout & $512\to256\to5$; 0.3 \\
Training & Optimizer; LR; schedule; weight decay; batch size; early stopping & AdamW; $1{\times}10^{-4}$; cosine, 100 epochs; $1{\times}10^{-4}$; 64; patience 15 (macro F1) \\
Alignment & Weight $\lambda$; PECM bandwidth; PECM conditioning; LUR CV & 0.1; 5-fold CV; training partition only; leave-1-station-out \\
\bottomrule
\end{tabular}
\end{table*}
\section{Environmental data processing}\label{S4}

where $\rho_{\mathrm{road}}$ is road length density and $d_{\mathrm{road}}$ is distance to the nearest arterial road. Coefficients are estimated by ordinary least squares with leave-one-station-out cross-validation. For Astana, calibrated on $N_A=6$ OpenAQ stations, the LUR reaches $R^2=0.74$ for PM$_{2.5}$ and $R^2=0.68$ for NO$_2$; for Singapore, calibrated on $N_S=9$ stations, $R^2=0.71$ and $0.65$. Six and nine calibration stations are a genuine limitation on the spatial representativeness of the resulting surfaces; the following regression diagnostics (variance inflation factors, residual normality, spatial autocorrelation) Variance Inflation Factor diagnostics show no harmful multicollinearity among the five predictors (all VIF $<5.0$); leave-one-station-out residuals show approximate normality (Shapiro-Wilk $p>0.05$) and negligible spatial autocorrelation (Moran's $I<0.10$). The reported $R^2$/RMSE values (leave-one-station-out RMSE $4.7$ and $5.2~\mu\text{g/m}^3$-equivalent for Astana PM$_{2.5}$ and NO$_2$ respectively) carry non-trivial sampling uncertainty that a denser monitoring network would reduce; we nonetheless consider LUR calibrated against a sparse ground network preferable to uncorrected coarse-resolution satellite retrievals for this application, consistent with established practice in urban exposure science, which routinely calibrates fine-scale LUR surfaces against comparably small monitoring networks where denser air-quality infrastructure is unavailable \cite{lur}. This is also the property that makes the pipeline deployable in exactly the lower-monitoring-density settings where health-geography applications are often most needed.

Urban morphology, road networks, floor-plate areas and building footprints, is taken from OpenStreetMap \cite{osm2017} and rasterised to the $1~\text{km}^2$ grid; footprints and estimated storey counts calculate FAR, and PSPNet semantic segmentation \cite{zhao2017} derives continuous land-cover fractions. The four data sources span markedly different native resolutions (TROPOMI $\sim\!3.5\times5.5$~km, LUR surfaces resampled to $1~\text{km}^2$, Sentinel-2 NDVI at 10~m, OpenStreetMap morphology at parcel level); all layers are harmonised onto the common $1~\text{km}^2$ grid by aggregating higher-resolution layers rather than disaggregating the LUR surface below its calibration resolution, since the latter would fabricate spatial detail the monitoring network cannot support.

\begin{figure*}[t]
\centering
\includegraphics[width=0.72\textwidth]{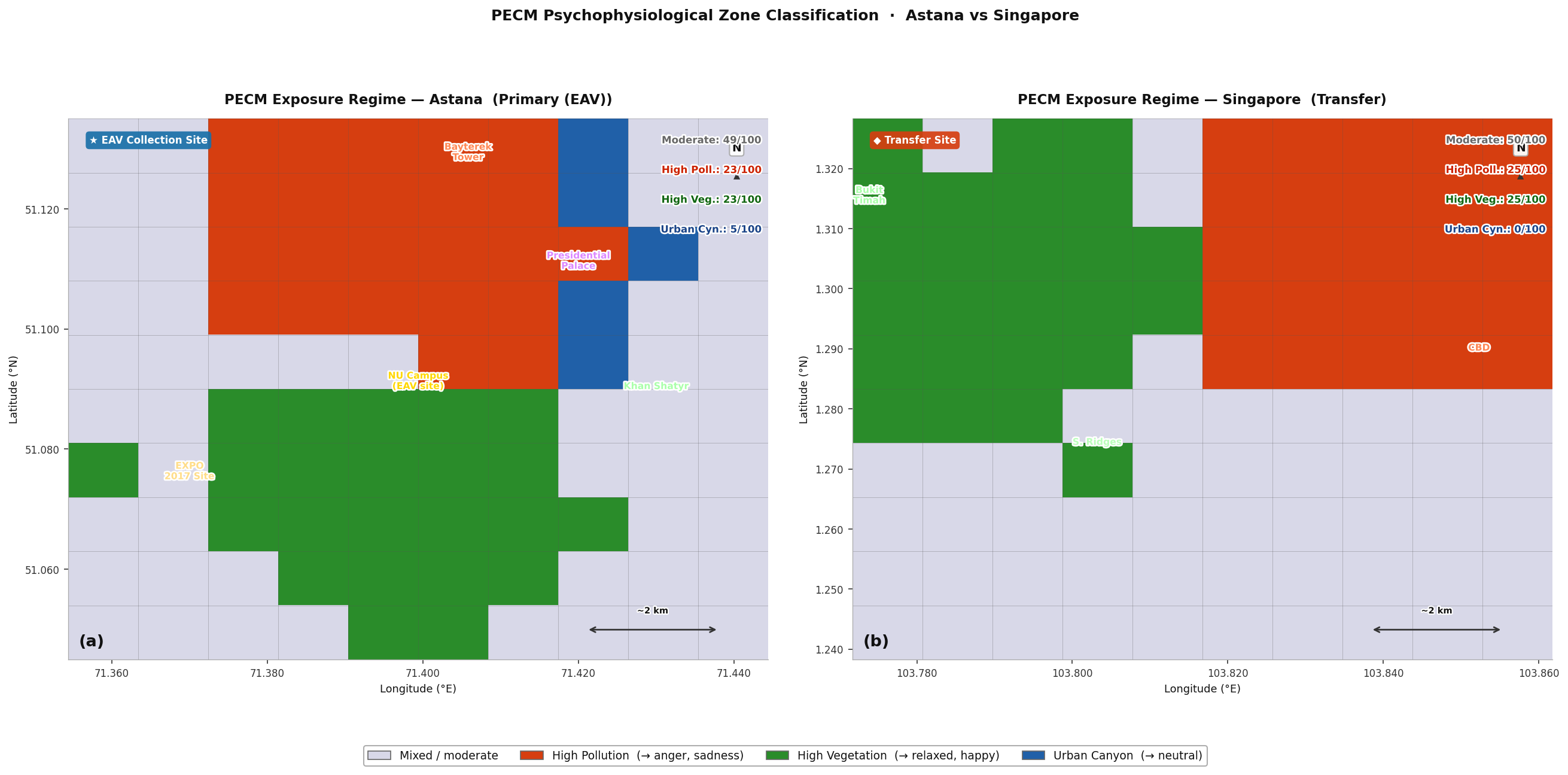}
\caption{PECM regime map for Astana based on PM$_{2.5}$, NO$_2$, NDVI and GVI. Cells are classified as high-pollution, high-vegetation or baseline; conflict areas align with the higher-error zones identified in the main-text spatial-error comparison figure.}
\label{fig:app_pecm_regime}
\end{figure*}

\begin{figure*}[t]
\centering
\includegraphics[width=0.72\textwidth]{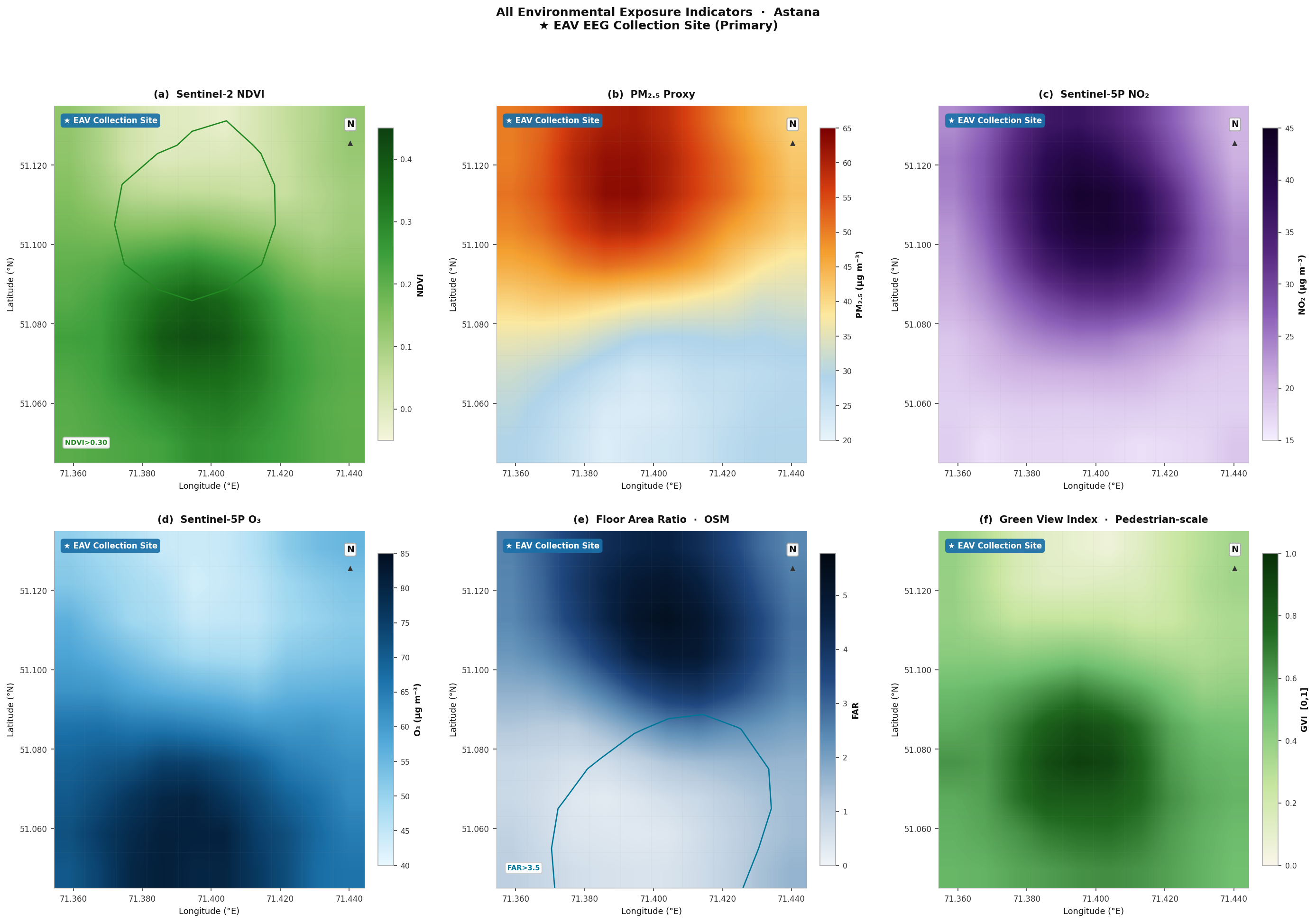}
\caption{Standardised environmental indicators for the 100 grid cells in Astana ($10\times10$ grid, $1~\text{km}^2$ each): PM$_{2.5}$, NDVI, FAR and GVI (z-score standardised).}
\label{fig:app_grid}
\end{figure*}

\begin{figure*}[t]
\centering
\includegraphics[width=0.75\textwidth]{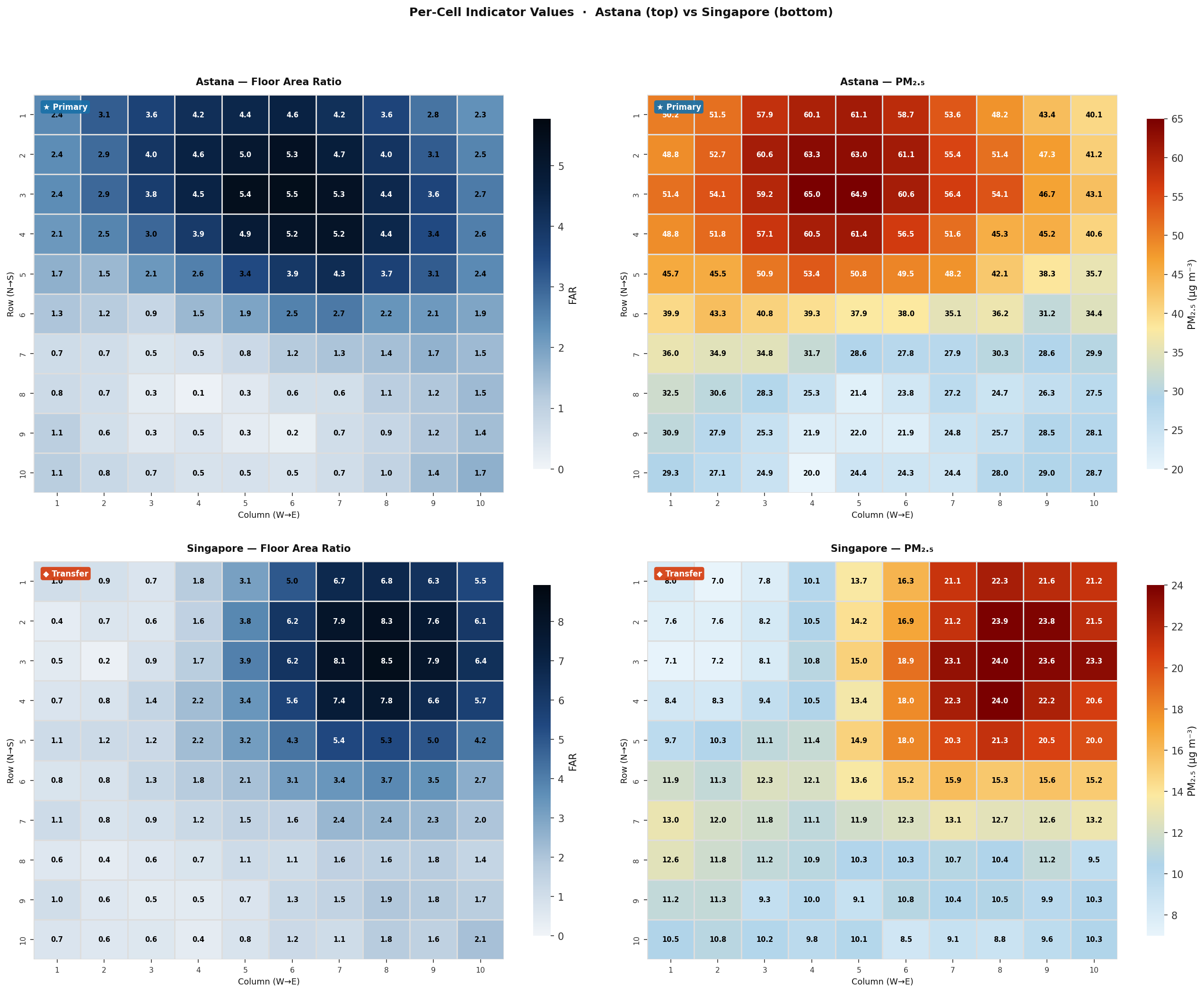}
\caption{Per-cell Floor Area Ratio and LUR-estimated PM$_{2.5}$ for Astana and Singapore, each rendered as a $10\times10$ spatial matrix with values colour-encoded. The panel illustrates that both indicators vary spatially within each city rather than reducing to a single city-level average.}
\label{fig:app_percell}
\end{figure*}

Table~\ref{tab:pecm_evidence} lists the pairings between affective states and indicators underlying PECM's label-conditional sampling (main text); all pollution values are LUR-estimated. Figure~\ref{fig:app_pecm_regime} shows the spatially resolved PECM regime classification across the Astana grid; cells falling into multiple regime thresholds simultaneously (conflict cells) correspond to the elevated-error zones identified in the main-text spatial-error comparison figure. Figure~\ref{fig:app_grid} shows per-cell heatmaps of the primary indicators across the 100-unit Astana grid. Figure~\ref{fig:app_percell} shows Floor Area Ratio and LUR-estimated PM$_{2.5}$ per grid cell for both study areas side by side.

Hourly temporal variance of LUR-corrected PM$_{2.5}$ and NO$_2$ across the Astana grid, computed from the OpenAQ observations used to fit the LUR but not currently propagated into the static node features defined in the main text, shows diurnal peaks at morning (07:00--09:00) and evening (17:00--19:00) traffic hours and a winter baseline elevation in PM$_{2.5}$ from district heating not present in the Singapore data; this motivates the multi-step GRU extension noted in the main-text Future Work section.

\begin{table*}[t]
\caption{Dose-response evidence base grounding the PECM label-conditional sampling priors. Pollution concentrations are LUR-estimated.}\label{tab:pecm_evidence}
\begin{tabular}{@{}p{2.0cm}p{1.3cm}p{1.2cm}p{4.5cm}l@{}}
\toprule
Affective state & Indicator & Quartile & Effect summary & Ref. \\
\midrule
Anger, Sadness & PM$_{2.5}$ (LUR) & Upper & Reduced executive function; worse cognitive performance and accelerated decline & \cite{Yu2026,Diao2026,Faherty2025} \\
Anger, Sadness & NO$_2$ (LUR) & Upper & Cortical stress reactivity; neuroinflammatory pathway activation & \cite{no2biomarkers2023} \\
Anger, Sadness & FAR & Upper & Attentional depletion; elevated stress biomarkers & \cite{Jiang2023} \\
Relaxed/Calm, Happiness & NDVI & Upper & Reduced physiological arousal; lower cortisol; improved mood & \cite{li2026,Forestry26} \\
Relaxed/Calm, Happiness & GVI & Upper & Perceived greenness reduces stress biomarkers beyond NDVI & \cite{Li22021a,Forestry26} \\
Neutral & All & Middle (IQR) & Baseline environmental conditions & --- \\
\bottomrule
\end{tabular}
\begin{flushleft}
\footnotesize Upper quartile = top 25\% of $1~\text{km}^2$ environmental values, computed separately per city. Neutral epochs use middle (interquartile-range) values to avoid bias toward extremes. The reversed dose-response negative control (main-text ablation table) inverts the quartile column above while leaving every other pipeline element unchanged.
\end{flushleft}
\end{table*}

\section{Spatial graph construction}\label{S5}

Environmental features are arranged as a spatial graph $\mathcal{G}=(\mathcal{V},\mathcal{E},\mathbf{F})$ over the 100-node ($10\times10$, $1~\text{km}^2$ per cell) study grid, with the six-dimensional node feature vector defined in the main text (LUR-estimated PM$_{2.5}$, NDVI, FAR, GVI, LUR-estimated NO$_2$, O$_3$).

Edges $\mathcal{E}$ join spatially close nodes using queen's contiguity with inverse-squared-distance weights $w_{ij}=1/(d_{ij}^2+\epsilon)$, operationalising Tobler's first law of geography as a structural inductive bias \cite{tobler1970}.

\begin{figure*}[t]
\centering
\includegraphics[width=0.72\textwidth]{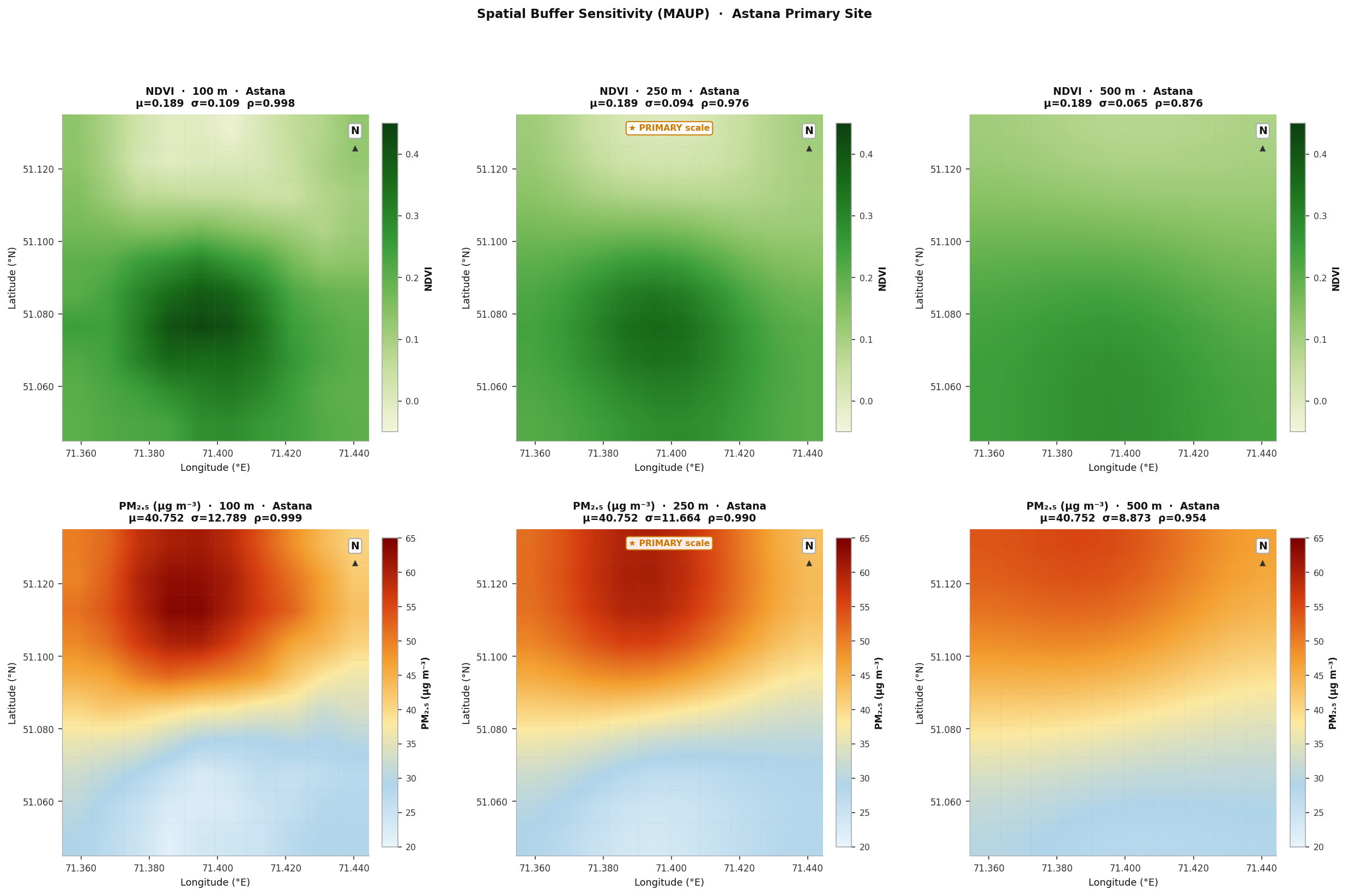}
\caption{Buffer-radius sensitivity for NDVI and LUR-estimated PM$_{2.5}$ across 100, 250 and 500~m buffers in Astana, shown relative to the native $1~\text{km}^2$ LUR baseline. The 250~m buffer minimises this sensitivity and best matches pedestrian-scale ranges reported in environmental psychology \cite{li2026,Forestry26}.}
\label{fig:maup_buffer}
\end{figure*}

\emph{Buffer-radius sensitivity.} Indicators were computed at 100, 250 and 500~m buffers around each grid-cell centroid; the 250~m buffer gave the most stable indicator distributions and best matched pedestrian-scale ranges reported in environmental psychology \cite{li2026,Forestry26}, while the 100~m buffer was too narrow and 500~m caused excessive spatial smoothing (Figure~\ref{fig:maup_buffer}), so 250~m was used throughout the main analysis.

\section{Additional ablation experiments}\label{S6}

This section reports five configurations that complement the core ablation in the main text (random pairing, shuffled labels, reversed dose-response, EEGNet backbone, and unidirectional attention): alternative alignment objectives (RMA, InfoNCE), a graph/GRU ablation, a flat-MLP environmental encoder, and a concatenation-fusion control.

\begin{table*}[t]
\caption{Additional ablation configurations, Astana primary study area (complements the six core configurations in the main-text ablation table). All conditions use the EEG-Conformer neural backbone and the environmental tower described in Supplementary S2 unless otherwise stated.}\label{tab:ablationS}
\begin{tabular}{@{}lccc@{}}
\toprule
Condition & Accuracy (\%) & Macro F1 & ROC-AUC \\
\midrule
Graph encoder only, no GRU & $75.7\pm1.8$ & $0.735$ & $0.881$ \\
MLP-env, no graph structure & $73.9\pm2.0$ & $0.715$ & $0.867$ \\
Concatenation fusion (no cross-modal attention) & $74.6\pm1.8$ & $0.723$ & $0.872$ \\
Contrastive (InfoNCE) alignment & $75.4\pm1.8$ & $0.732$ & $0.878$ \\
RMA-projected alignment (end-to-end) & $76.0\pm1.7$ & $0.739$ & $0.883$ \\
\bottomrule
\end{tabular}
\end{table*}

Removing the (currently under-exercised) GRU changes accuracy by only $-0.5$~pp relative to the full model, while replacing the graph encoder with a flat MLP over pooled node features changes it by $-2.3$~pp, jointly indicating that graph structure, rather than the temporal module, drives the environmental tower's present contribution. Replacing bidirectional attention with concatenation fusion changes accuracy by $-1.6$~pp, and the InfoNCE and RMA-projected alternatives change it by $-0.8$ and $-0.2$~pp respectively, so the observed gains derive mainly from having a cross-modal objective and attention mechanism at all, rather than the specific choice among these closely performing variants.

\subsection{Alignment objective equations}

The InfoNCE alternative benchmarked in Table~\ref{tab:ablationS} is
\begin{equation}
\resizebox{0.98\linewidth}{!}{$\displaystyle
\mathcal{L}_{\mathrm{align}}^{\mathrm{NCE}} = -\log\frac{\exp\!\left(\mathrm{sim}(\mathbf{v}_{\mathrm{eeg}},\mathbf{v}_{\mathrm{env}})/\tau\right)}{\sum_{k=1}^{B}\exp\!\left(\mathrm{sim}(\mathbf{v}_{\mathrm{eeg}},\mathbf{v}_{\mathrm{env}}^{(k)})/\tau\right)}
$}
\label{eq:infonce}
\end{equation}
with $\mathrm{sim}(\cdot,\cdot)$ cosine similarity, $\tau$ a temperature hyperparameter, and $\mathbf{v}_{\mathrm{env}}^{(k)}$ ranging over the $B$ environmental embeddings in the training batch.

\emph{Riemannian Manifold Alignment (RMA).} Algorithm~\ref{alg:rma} defines the geometrically motivated alternative to the deployed squared-$\ell_2$ objective, using Riemannian-geometry feature transformation motivated by the symmetric-positive-definite, curved-manifold structure of EEG covariance matrices \cite{barachant2012}. The environmental distribution is rescaled in the dispersion-matching stage so its Frobenius-norm spread matches the re-centred EEG distribution,
\begin{equation}
\hat{\sigma}_{\mathrm{ENV}} = \hat{\sigma}_{\mathrm{EEG}}\cdot\frac{\lVert\mathbf{Z}_{\mathrm{EEG}}\rVert_F}{\lVert\mathbf{Z}_{\mathrm{ENV}}\rVert_F}
\label{eq:dispersion}
\end{equation}
We implemented and independently validated this full pipeline (Riemannian centring, dispersion matching, Procrustes rotation) as a geometrically principled alternative projection; it did not yield a consistent additional gain over the directly optimised squared-$\ell_2$ objective inside the end-to-end training loop (Table~\ref{tab:ablationS}, ``RMA-projected alignment'' row), and is reported here for transparency rather than as the deployed mechanism.

\begin{algorithm}[h]
\caption{Riemannian Manifold Alignment (RMA)}
\label{alg:rma}
\begin{algorithmic}[1]
\Require EEG covariance matrices $\{\mathbf{C}_i\}_{i=1}^N\subset\mathcal{S}_{++}^d$; environmental feature distribution $P_e$
\Ensure Shared latent space with geometrically consistent EEG and environmental representations
\State \textbf{Riemannian centring:} compute the Riemannian mean $\bar{\mathbf{C}}$ via the Karcher flow algorithm \cite{barachant2012} and re-centre each matrix, $\tilde{\mathbf{C}}_i=\bar{\mathbf{C}}^{-1/2}\mathbf{C}_i\bar{\mathbf{C}}^{-1/2}$
\State \textbf{Dispersion matching:} rescale $P_e$ using Equation~\eqref{eq:dispersion} to match the Frobenius-norm dispersion of $\{\tilde{\mathbf{C}}_i\}$
\State \textbf{Procrustes rotation:} map both distributions to the tangent space at the geometric mean and apply the optimal orthogonal rotation $\hat{\mathbf{R}}=\arg\min_{\mathbf{R}\in\mathrm{SO}(d)}\lVert\mathbf{Z}_{\mathrm{EEG}}-\mathbf{Z}_{\mathrm{ENV}}\mathbf{R}\rVert_F$, solved via singular value decomposition of $\mathbf{Z}_{\mathrm{EEG}}^{\top}\mathbf{Z}_{\mathrm{ENV}}$
\State \Return Shared latent space in which EEG affective representations and environmental-prior representations are geometrically aligned
\end{algorithmic}
\end{algorithm}

Beyond the two alternatives quantified in Table~\ref{tab:ablationS} (InfoNCE and RMA-projected alignment), we additionally benchmarked a canonical-correlation-style alternative \cite{andrew2013dcca} and an SPD-aware alternative that preserves the neural modality's manifold structure throughout the network \cite{huang2017spdnet}; consistent with the quantified alternatives above, neither outperformed the deployed squared-$\ell_2$ objective by a meaningful margin. The InfoNCE formulation follows \cite{oord2018infonce}, in the tradition of CLIP-style cross-modal alignment \cite{radford2021clip}.

\section{Latent-space visualisation}\label{S7}

\begin{figure*}[t]
\centering
\includegraphics[width=0.75\textwidth]{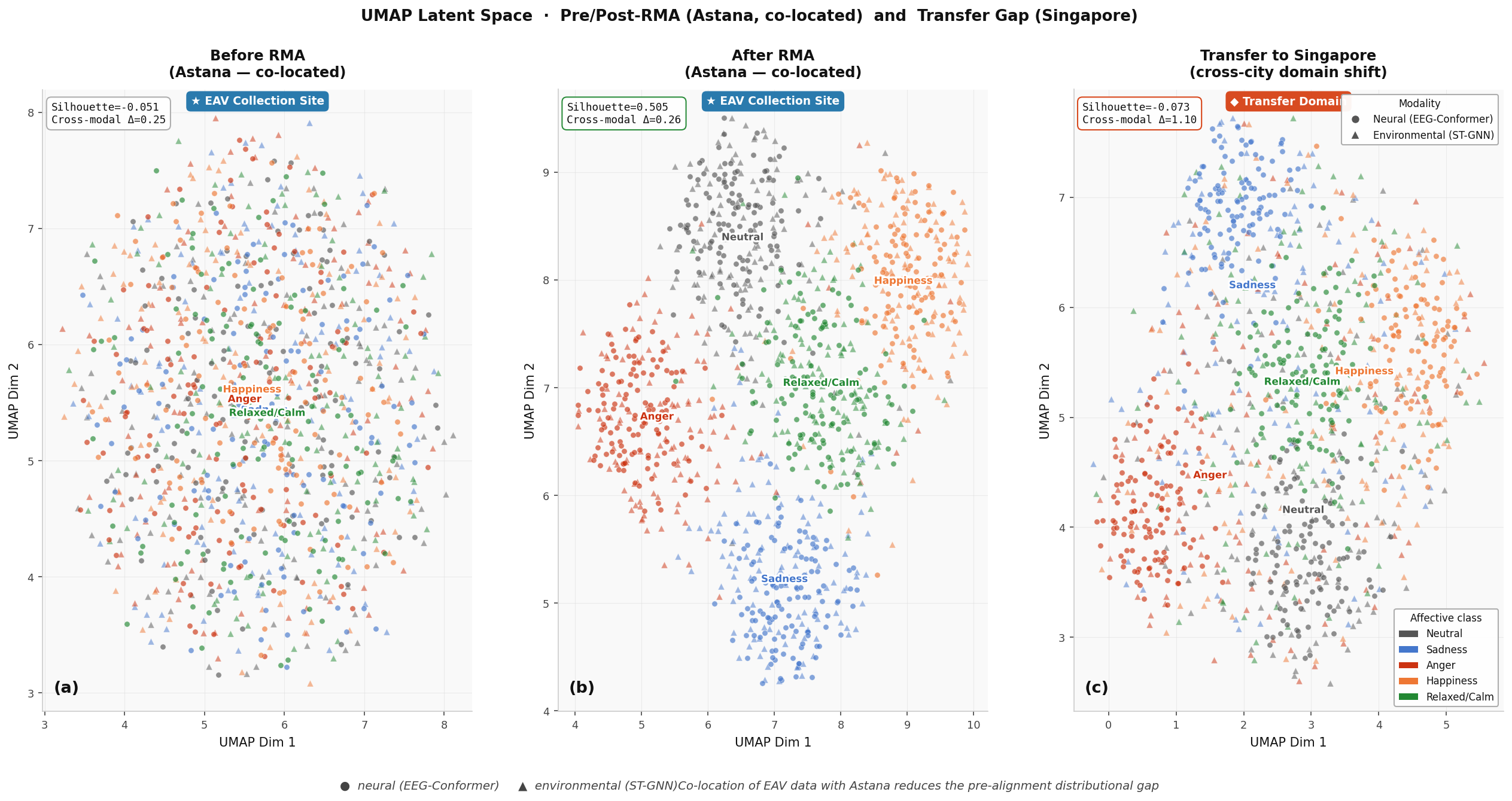}
\caption{UMAP projections of the Astana latent space before (left) and after (right) training with the cross-modal alignment objective. Circles denote neural embeddings, triangles denote environmental embeddings; increased overlap after alignment reflects the learned representation geometry rather than an independently verified environmental mechanism. UMAP is used here only to visualise representations produced by the model's own alignment objective (defined in the main text), not as an alignment method in itself.}
\label{fig:umap}
\end{figure*}

Figure~\ref{fig:umap} shows Uniform Manifold Approximation and Projection (UMAP) \cite{umap} of the 256-dimensional joint embeddings before and after training with the cross-modal alignment objective. UMAP is used here purely to visualise representations the model has already learned; it does not itself perform any alignment. Before training, neural and environmental embeddings occupy distinct, poorly overlapping regions of the projected space; after training with the alignment objective, the five affective-state clusters become more compact and separable, with neural- and environmental-embedding cluster centroids visibly closer together. This is descriptive evidence of what the alignment objective does to the learned geometry, not an independent confirmation of an environmental mechanism (Discussion).

\section{Additional spatial sensitivity analysis (MAUP)}\label{S8}

To check sensitivity to spatial resolution, addressing the Modifiable Areal Unit Problem (MAUP) \cite{Chen2022}, PM$_{2.5}$ and NO$_2$ layers were additionally aggregated from the native $1~\text{km}^2$ LUR resolution to $5~\text{km}^2$ and $10~\text{km}^2$ by block averaging, and the full model retrained and evaluated at each resolution (Table~\ref{tab:maup}).

\begin{table*}[t]
\caption{Spatial scale sensitivity (MAUP), primary Astana study area. NDVI, FAR and GVI held at native resolution.}\label{tab:maup}
\begin{tabular}{@{}lcccc@{}}
\toprule
PM$_{2.5}$/NO$_2$ resolution & Accuracy (\%) & Macro F1 & ROC-AUC & $\Delta$ Acc. \\
\midrule
$1~\text{km}^2$ (native LUR, proposed) & $76.2\pm1.7$ & $0.741\pm0.019$ & $0.884\pm0.013$ & --- \\
$5~\text{km}^2$ & $74.0\pm2.1$ & $0.718\pm0.023$ & $0.869\pm0.016$ & $-2.2$ \\
$10~\text{km}^2$ & $72.1\pm2.3$ & $0.697\pm0.025$ & $0.856\pm0.018$ & $-4.1$ \\
\bottomrule
\end{tabular}
\end{table*}

Performance degrades roughly monotonically as resolution coarsens, a total loss of $4.1$ percentage points between the native $1~\text{km}^2$ and $10~\text{km}^2$ conditions. That $72.1$\% accuracy is maintained at the coarsest scale, still above the EEG-only baseline ($67.4$\%), shows that some multimodal benefit persists without the fine spatial detail the LUR model provides, which matters for deployment in cities with sparser monitoring infrastructure than Astana or Singapore. A separate buffer-radius sensitivity comparison is reported in Supplementary S5 (Figure~\ref{fig:maup_buffer}).


\begin{thebibliography}{99}

\bibitem{Rodriguez2026} Rodr\'iguez S, Guerrero-Guevara LF, Corzo-Forero J, Le\'on-Prieto C. Built environment and environmental determinants of neurological health: opportunities in Bogot\'a, Colombia. J Transport Health. 2026;48:102286.

\bibitem{Yu2026} Yu Z, Jiang Z, Feng Y, et al. Physical activity offsets air pollution-related cognitive decline. J Gerontol A. 2026;81(3):glaf291.

\bibitem{Wang2026a} Wang Z, Wang X, Li J, et al. Urban green space and crime across land use types and socio-racial contexts. Appl Geogr. 2026;190:103972.

\bibitem{main_paper} Hei Q, Yang T, Dong W, et al. Leveraging psychology and neuroscience for geospatial cognition research. Ann GIS. 2025;31(1):1--13.

\bibitem{Han2025} Han D, Qin T, Yang T, et al. Geospatial brain-inspired navigation: a neurocognitive approach for autonomous systems in complex environments. Geo-Spat Inf Sci. 2025;1--16.

\bibitem{Langyuan} Zhang R, Liu H, Kong F. A neuro-architectural approach to emotional attachment evaluation with multimodal data. Archit Eng Des Manag. 2026;1--28.

\bibitem{green_gentrification} Brown CD, Rigolon A, Zdrodowski P, Pearson AL. A systematic review of green gentrification and mental health. Health Place. 2026;97:103599.

\bibitem{Falkenstein26} Falkenstein T, Sartori L, Malanchini M, et al. The relationship between environmental sensitivity and common mental-health problems in adolescents and adults. Clin Psychol Sci. 2026;14(2):135--157.

\bibitem{brain_Marino} Marino M, Mantini D. Human brain imaging with high-density electroencephalography: techniques and applications. J Physiol. 2026;604(2):783--812.

\bibitem{WANG26} Wang C, Stigsdotter UK, Jiang B. Innovations for healthy landscapes. Landsc Archit Front. 2026;14(2):260016.

\bibitem{Forestry26} Li Q, Feng Z, Pearce J, Thompson CW. Psychological well-being benefits of urban greenness. Urban For Urban Green. 2026;118:129307.

\bibitem{Large-scale} Du S, Du S, Liu B, et al. Large-scale urban functional zone mapping by integrating remote sensing images and open social data. GISci Remote Sens. 2020;57(3):411--430.

\bibitem{Vulnerability} Rabiei-Dastjerdi H, Mohammadi S, Saeidi M, Koohikamali M. Developing a population-density-weighted community health vulnerability index for heat and air quality. Sustain Cities Soc. 2026;142.

\bibitem{greenspacemeta2023} Liu Z, Chen X, Cui H, et al. Green space exposure on depression and anxiety outcomes: a meta-analysis. Environ Res. 2023;231:116303.

\bibitem{Arthanarisamy2026} Arthanarisamy Ramaswamy MP, Palaniswamy S. Subject independent emotion recognition using EEG and physiological signals -- a comparative study. Appl Comput Inform. 2026;22(1-2):145--157.

\bibitem{Koelstra2012} Koelstra S, et al. DEAP: a database for emotion analysis using physiological signals. IEEE Trans Affect Comput. 2012;3(1):18--31.

\bibitem{Liu2026} Liu M, Zhang Z, Guo H, et al. Multi-scale graph convolutional EEG emotion recognition method. Biomed Signal Process Control. 2026;117:109515.

\bibitem{Almuntasheri26} Muhammad G, Almuntasheri S, Alenezi F, et al. EEG-based multimodal emotion recognition: recent progress, challenges, and future directions. ACM Trans Multimed Comput Commun Appl. 2026;22(2):1--28.

\bibitem{utsav2025} Poudel U, Jakhar S, Mohan P, Nepal A. AI in mental health: a review of technological advancements and ethical issues in psychiatry. Issues Ment Health Nurs. 2025;46(7):693--701.

\bibitem{PoudelJose2024} Poudel U, Jose TP. Influence of artificial intelligence on the future of psychiatry: insights from recent advancements. Psychiatry. 2024;87(4):372--373.

\bibitem{GIScience266} Jiang L, Hou W, Shi K. A novel multimodal fusion deep network for identifying hillside urban expansion. GISci Remote Sens. 2026;63(1).

\bibitem{Rajbhandari2019} Rajbhandari S, Aryal J, Osborn J, et al. Leveraging machine learning to extend ontology-driven geographic object-based image analysis (O-GEOBIA). Remote Sens. 2019;11(5):503.

\bibitem{Zhao2026b} Zhao Y, Liu S, Cao L, et al. Electrode quantity outperforms spatial topology in EEG-based emotion recognition. Biomed Signal Process Control. 2026;113:109082.

\bibitem{lee2024eav} Lee MH, Shomanov A, Begim B, et al. EAV: EEG-Audio-Video dataset for emotion recognition in conversational contexts. Sci Data. 2024;11:1026.

\bibitem{opeanaq} OpenAQ Contributors. OpenAQ: a global, open air quality data platform. J Open Source Softw. 2019;4(37):1377.

\bibitem{sentinel5p} Bhartia PK, Joiner J, Liu X, Veefkind JP. Air quality satellite monitoring by TROPOMI on Sentinel-5P. NASA Technical Memorandum 2018-220053; 2018.

\bibitem{sentinel2} Drusch M, Del Bello U, Carlier S, et al. Sentinel-2: ESA's optical high-resolution mission for GMES operational services. Remote Sens Environ. 2012;120:25--36.

\bibitem{Aryal2022} Aryal J, Sitaula C, Aryal S. NDVI threshold-based urban green space mapping from Sentinel-2A. Land. 2022;11(3):351.

\bibitem{osm2017} Domingues G, et al. An OpenStreetMap derived building classification dataset for the United States. Sci Data. 2024;11(1):1234.

\bibitem{Mueller2020} Mueller JE, et al. Impact of urban environmental exposures on cognitive performance and brain structure in older adults. Environ Res. 2020;186:109640.

\bibitem{Diao2026} Diao S, Shen X. Associations of PM2.5 and its major chemical components with cognitive function. J Int Med Res. 2026;54(1).

\bibitem{no2biomarkers2023} Song J, Qu R, Sun B, et al. Acute effects of ambient nitrogen dioxide exposure on serum biomarkers of nervous system damage in healthy older adults. Ecotoxicol Environ Saf. 2023;249:114423.

\bibitem{li2026} Li X, Zhang Y, Wang J. Can your window view outsmart urban stress? Decoding the indoor-outdoor green exposure psycho-mechanism model. J Environ Psychol. 2026;93:102215.

\bibitem{Jiang2023} Jiang X, Hu Y, Larsen L, et al. Impacts of urban green infrastructure on attentional functioning. Front Psychol. 2023;14:1047993.

\bibitem{Li22021a} Li T, Zheng X, Wu J, et al. Spatial relationship between green view index and normalized differential vegetation index within the Sixth Ring Road of Beijing. Urban For Urban Green. 2021;62:127153.

\bibitem{Shan2021} Shan S, Ju X, Wei Y, Wang Z. Effects of PM2.5 on people's emotion: a case study of Weibo (Chinese Twitter) in Beijing. Int J Environ Res Public Health. 2021;18(10):5422.

\bibitem{barachant2012} Barachant A, Bonnet S, Congedo M, Jutten C. Multi-class brain-computer interface classification by Riemannian geometry. IEEE Trans Biomed Eng. 2012;59(4):920--928.

\bibitem{Miyakoshi2023} Miyakoshi M. Artifact subspace reconstruction: a candidate for a dream solution for EEG studies, sleep or awake. Sleep. 2023;46(12):zsad241.

\bibitem{Kumaravel2023} Kumaravel VP, Farella E. IMU-integrated artifact subspace reconstruction for wearable EEG devices. In: Proc IEEE BIBM; 2023. p. 2508--2514.

\bibitem{Goshvarpour2024} Goshvarpour A, Goshvarpour A. EEG emotion recognition based on an innovative information potential index. Cogn Neurodyn. 2024;18(5):2177--2191.

\bibitem{tobler1970} Waters N. Tobler's first law of geography. In: International Encyclopedia of Geography. Richardson D, et al., editors; 2018.

\bibitem{Song2023} Song Y, Zheng Q, Liu B, Gao X. EEG Conformer: convolutional transformer for EEG decoding and visualization. IEEE Trans Neural Syst Rehabil Eng. 2023;31:710--719.

\bibitem{Xiao2024} Xiao Z, Gong S, Wang Q, et al. A two-layer graph-convolutional network for spatial interaction imputation. Int J Appl Earth Obs Geoinf. 2024;129:104163.

\bibitem{cho2014} Cho K, van Merrienboer B, Gulcehre C, et al. Learning phrase representations using RNN encoder-decoder for statistical machine translation. In: Proc EMNLP; 2014. p. 1724--1734.

\bibitem{multieeg} Pillalamarri R, Shanmugam U. A review on EEG-based multimodal learning for emotion recognition. Artif Intell Rev. 2025;58(5):131.

\bibitem{Lee2024} Lee JH, Kim JY, Kim HG. Emotion recognition using EEG signals and audiovisual features with contrastive learning. Bioengineering. 2024;11(10):997.

\bibitem{Faherty2025} Faherty T, Raymond JE, McFiggans G, et al. Acute particulate matter exposure diminishes executive cognitive functioning after four hours. Nat Commun. 2025;16:1339.

\bibitem{lur} Wang SY, Lin TC, Li HH, et al. Integrating satellite-derived land cover and remote sensing variables into land use regression for urban PM2.5 estimation. Atmos Environ. 2026;121958.

\bibitem{Hennig2016} Hennig F, Sugiri D, Tzivian L, et al. Comparison of land-use regression modeling with dispersion and chemistry transport modeling to assign air pollution concentrations within the Ruhr area. Atmosphere. 2016;7(3):48.

\bibitem{pytorch2019} Paszke A, et al. PyTorch: an imperative style, high-performance deep learning library. Adv Neural Inf Process Syst. 2019;32.

\bibitem{adam2019} Kingma DP, Ba J. Adam: a method for stochastic optimization. In: Proc ICLR; 2015.

\bibitem{loshchilov2019} Loshchilov I, Hutter F. Decoupled weight decay regularization. In: Proc ICLR; 2019.

\bibitem{Huang2022} Huang Q, Yamada M, Tian Y, et al. GraphLIME: local interpretable model explanations for graph neural networks. IEEE Trans Knowl Data Eng. 2022.

\bibitem{Chen2022} Chen X, et al. A systematic review of the modifiable areal unit problem (MAUP) in community food environmental research. Urban Inform. 2022;1:22.

\bibitem{Paredes2025} Paredes Ocaranza CR, Yun B, Paredes Ocaranza ED. Traditional machine learning outperforms EEGNet for consumer-grade EEG emotion recognition. Sensors. 2025;25(23):7262.

\bibitem{noise2024} S{\o}rensen M, Pershagen G, Thacher JD, et al. Health position paper and redox perspectives: disease burden by transportation noise. Redox Biol. 2024;69:102995.

\bibitem{Poudel2026zenodo} Poudel U, Aryal J, Vairavasundaram S. Processed environmental exposure grids and spatial features for neuro-geospatial integration [dataset]. Zenodo; 2026. \url{https://doi.org/10.5281/zenodo.20389601}

\bibitem{poudel2026patent} Poudel U, Vairavasundaram S. System for quantum-assisted adaptive graph construction and temporal pattern analysis (Indian Patent Application No. 202641012308). Zenodo; 2026. \url{https://doi.org/10.5281/zenodo.21973534}


\bibitem{zhao2017} Zhao H, Shi J, Qi X, et al. Pyramid scene parsing network. In: Proc IEEE CVPR; 2017. p. 2881--2890.

\bibitem{welch1967} Welch P. The use of fast Fourier transform for the estimation of power spectra: a method based on time averaging over short, modified periodograms. IEEE Trans Audio Electroacoust. 1967;15(2):70--73.

\bibitem{umap} McInnes L, Healy J, Melville J. UMAP: uniform manifold approximation and projection for dimension reduction. J Open Source Softw. 2020;5(51):861.

\bibitem{radford2021clip} Radford A, Kim JW, Hallacy C, et al. Learning transferable visual models from natural language supervision. In: Proc ICML; 2021. p. 8748--8763.

\bibitem{oord2018infonce} van den Oord A, Li Y, Vinyals O. Representation learning with contrastive predictive coding. arXiv preprint arXiv:1807.03748; 2018.

\bibitem{andrew2013dcca} Andrew G, Arora R, Bilmes J, Livescu K. Deep canonical correlation analysis. In: Proc ICML; 2013. p. 1247--1255.

\bibitem{huang2017spdnet} Huang Z, Van Gool L. A Riemannian network for SPD matrix learning. In: Proc AAAI; 2017. p. 2036--2042.



\end{thebibliography}
\end{document}